\documentclass[10pt,twocolumn,letterpaper]{article}

\usepackage[pagenumbers]{cvpr} %

\usepackage[T1]{fontenc}    %
\usepackage{url}            %
\usepackage{booktabs}       %
\usepackage{amsfonts}       %
\usepackage{nicefrac}       %
\usepackage{microtype}      %
\usepackage{xcolor}         %
\usepackage{wrapfig}
\usepackage{balance}
\usepackage{titletoc}
\usepackage{graphicx}

\usepackage{amsthm}
\usepackage{amsmath}
\usepackage{amsfonts}
\usepackage{amssymb}
\usepackage{caption}
\usepackage{subcaption}
\usepackage{dsfont}
\usepackage{mathtools}
\usepackage{bbm}
\usepackage{bm}
\usepackage{multirow}
\usepackage{multicol}

\usepackage[symbol]{footmisc}
\usepackage{float}

\newcommand{\Reals}{\mathbb{R}}

\newcommand{\Lagrangian}{\mathcal{L}}

\newcommand{\veryshortarrow}[1][3pt]{\mathrel{%
   \vcenter{\hbox{\rule[-.5\fontdimen8\textfont3]{#1}{\fontdimen8\textfont3}}}%
   \mkern-4mu\hbox{\usefont{U}{lasy}{m}{n}\symbol{41}}}}

\DeclareMathOperator*{\argmax}{arg\,max}
\DeclareMathOperator*{\argmin}{arg\,min}

\let\emptyset\varnothing

\theoremstyle{definition}
\newtheorem{definition}{Definition}[section]

\definecolor{blue}{RGB}{66, 133, 244}
\definecolor{red}{RGB}{219, 68, 55}
\definecolor{yellow}{RGB}{244, 188, 0}
\definecolor{green}{RGB}{15, 157, 88}
\definecolor{dark}{RGB}{31, 31, 31}
\definecolor{purple}{RGB}{113, 78, 163}
\definecolor{orange}{RGB}{187, 85, 39}
\definecolor{indigo}{RGB}{63, 81, 181}

\definecolor{metalgreen}{RGB}{51, 157, 144}
\definecolor{metalorange}{RGB}{244, 161, 97}

\usepackage[pagebackref,breaklinks,colorlinks]{hyperref}
\hypersetup{
    colorlinks,
    linkcolor={indigo},
    citecolor={indigo},
    urlcolor={indigo},
    anchorcolor={indigo}
}

\newcommand{\pred}{\bm{f}}
\newcommand{\vx}{\mathbf{X}}
\newcommand{\vy}{\mathbf{Y}}
\newcommand{\vu}{\mathbf{U}}
\newcommand{\vw}{\mathbf{W}}
\newcommand{\vh}{\bm{h}}
\newcommand{\vg}{\bm{g}}
\newcommand{\va}{\mathbf{A}}
\newcommand{\vm}{\mathbf{M}}
\newcommand{\Var}{\mathbb{V}}

\newcommand{\metric}{\textit{Utility}}

\usepackage[capitalize]{cleveref}
\crefname{section}{Sec.}{Secs.}
\Crefname{section}{Section}{Sections}
\Crefname{table}{Table}{Tables}
\crefname{table}{Tab.}{Tabs.}

\def\cvprPaperID{5667} %
\def\confName{CVPR}
\def\confYear{2023}

\begin{document}

\title{CRAFT: Concept Recursive Activation FacTorization for Explainability
\vspace{-5mm}
}

\author{
    \hspace{0cm} 
    \textbf{Thomas Fel}$^{1,3,5}$\footnotemark[1]
    \hspace{1mm}
    \textbf{Agustin Picard}$^{3,6}$\footnotemark[1]
    \hspace{1mm}
    \vspace{1mm}
    \textbf{Louis Bethune}$^{3}$\footnotemark[1] 
    \hspace{0cm}
    \textbf{Thibaut Boissin}$^{3,4}$\footnotemark[1]
    \\
    \hspace{0cm}
    \textbf{David Vigouroux}$^{3,4}$
    \hspace{0cm}
    \textbf{Julien Colin}$^{1,3}$
    \hspace{0cm}
    \vspace{1mm}
    \textbf{Rémi Cadène}$^{1,2}$
    \hspace{0cm}
    \textbf{Thomas Serre}$^{1,3}$
    \vspace{1mm}
    \vspace{0.0cm}\\
{\normalsize 
$^1$Carney Institute for Brain Science, Brown University, USA \hspace{0.0cm} $^2$Sorbonne Université, CNRS, France}
\\
\vspace{-0mm}
{\normalsize $^3$Artificial and Natural Intelligence Toulouse Institute, Université de Toulouse, France}
\\
\vspace{-0mm}
{\normalsize $^4$ Institut de Recherche Technologique Saint-Exupery, France}
\\
\vspace{-0mm}
{\normalsize $^5$ Innovation \& Research Division, SNCF ~, $^6$ Scalian}
\vspace{1mm}
}

\twocolumn[{%
\renewcommand\twocolumn[1][]{#1}%
\maketitle
\vspace{-14mm}
\begin{minipage}{\textwidth}
\begin{figure}[H]\centering
\includegraphics[width=0.99\textwidth]{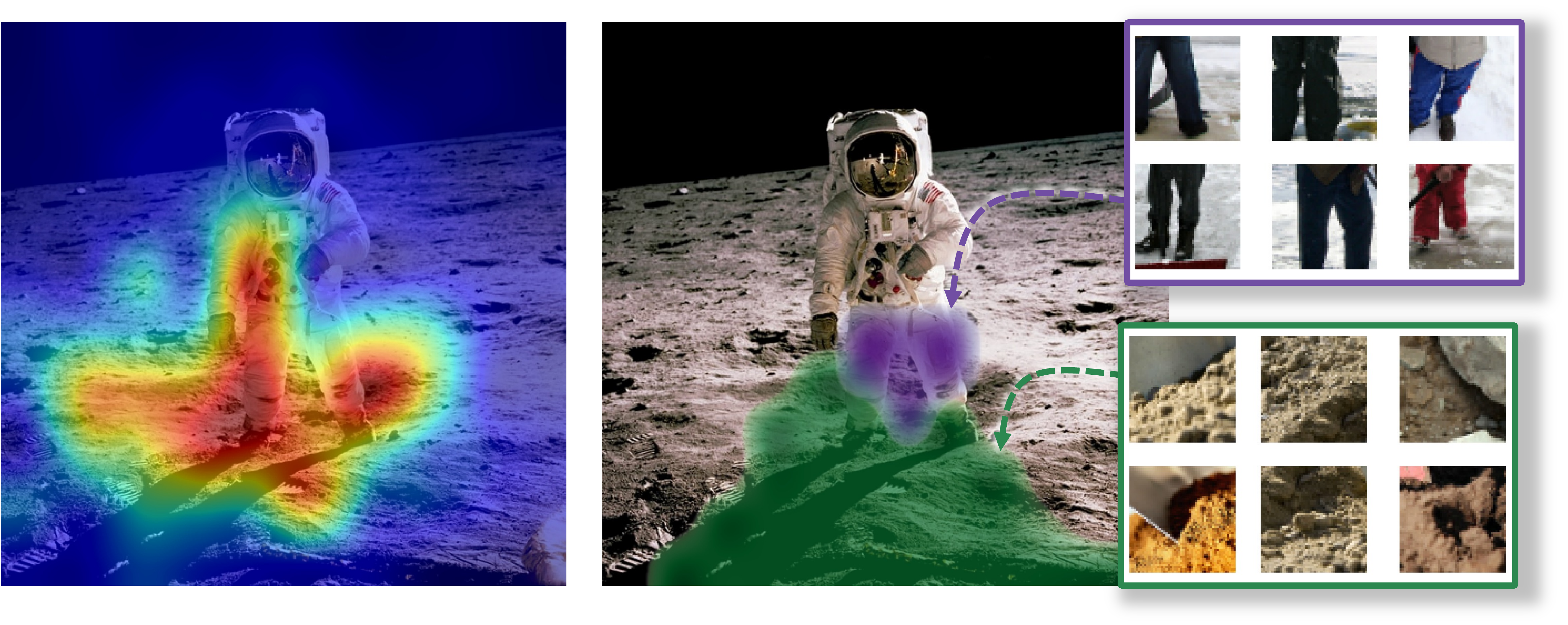}
\vspace{-2mm}
\caption{\textbf{The ``Man on the Moon'' incorrectly classified as a ``shovel'' by an ImageNet-trained ResNet50.} Heatmap generated by a classic attribution method~\cite{RISE} (left) vs.  \textit{concept attribution maps} generated with the proposed CRAFT approach (right) which highlights the two most influential concepts that drove the ResNet50's decision along with their corresponding locations. 
 CRAFT suggests that the neural net arrived at its decision because it identified the concept of ``dirt'' \textcolor{green}{$\bullet$} commonly found in members of the image class ``shovel'' and the concept of ``ski pants'' \textcolor{purple}{$\bullet$} typically worn by people clearing snow from their driveway with a shovel instead the correct concept of astronaut's pants (which was probably never seen during training).
  \vspace{3mm}
}
\label{fig:shovel}
\end{figure}
\end{minipage}
}]

\begin{abstract}

\vspace{-4mm}

Attribution methods, which employ heatmaps to identify the most influential regions of an image that impact model decisions, have gained widespread popularity as a type of explainability method.
However, recent research has exposed the limited practical value of these methods, attributed in part to their narrow focus on the most prominent regions of an image -- revealing "where" the model looks, but failing to elucidate "what" the model sees in those areas.
In this work, we try to fill in this gap with CRAFT -- a novel approach to identify both ``what'' and ``where'' by generating concept-based explanations.
We introduce 3 new ingredients to the automatic concept extraction literature: (\textbf{i}) a recursive strategy to detect and decompose concepts across layers, (\textbf{ii}) a novel method for a more faithful estimation of concept importance using Sobol indices, and (\textbf{iii}) the use of implicit differentiation to unlock Concept Attribution Maps.

We conduct both human and computer vision experiments to demonstrate the benefits of the proposed approach. We show that the proposed concept importance estimation technique is more faithful to the model than previous methods. When evaluating the usefulness of the method for human experimenters on a human-centered utility benchmark, we find that our approach significantly improves on two of the three test scenarios. Our code is freely available:
\href{https://github.com/deel-ai/Craft}{\nolinkurl{github.com/deel-ai/Craft}}

\end{abstract}

\vspace{-5.5mm}
\section{Introduction}
\vspace{-1mm}

\let\thefootnote\relax\footnotetext{\hspace{-6mm}* Equal contribution \\ \textit{Proceedings of the IEEE / CVF Computer Vision and Pattern Recognition Conference (CVPR), 2023.}}

\begin{figure*}[t!]\centering
  \includegraphics[width=0.75\textwidth]{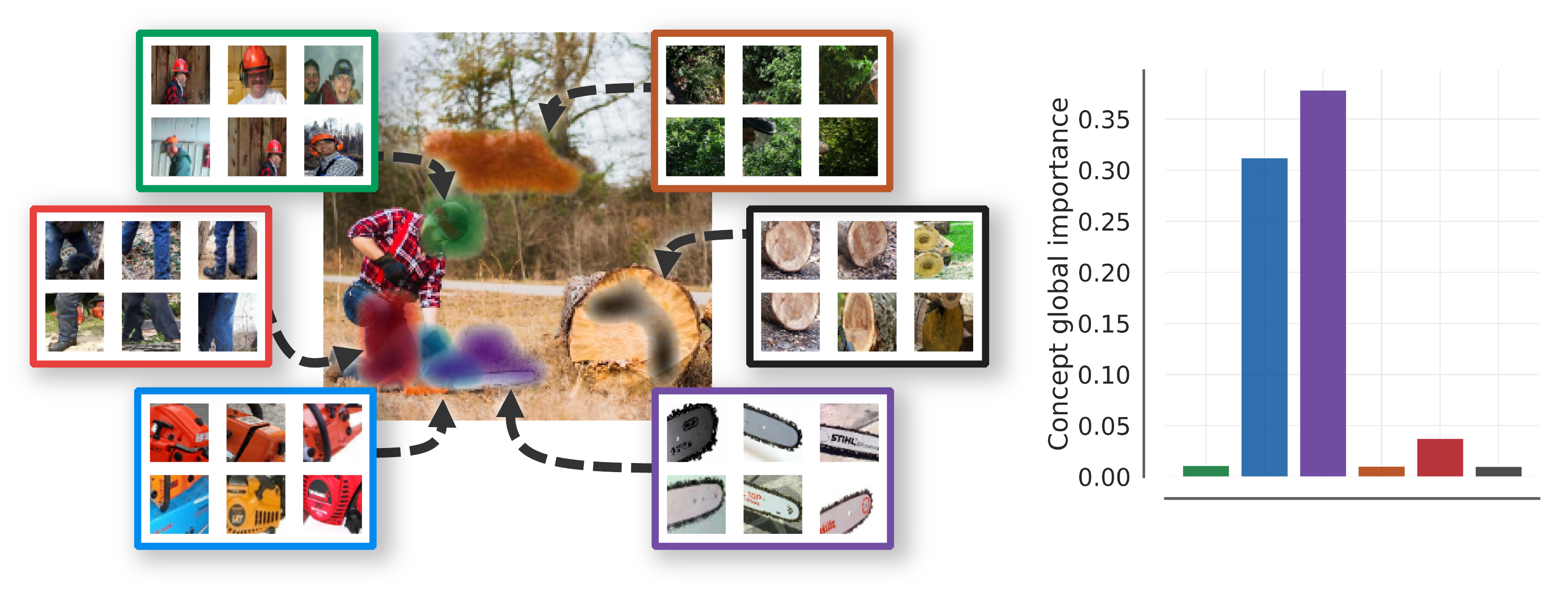}
  
  \vspace{-4mm}
  
\caption{\textbf{CRAFT results for the prediction ``chain saw''.} 
First, our method uses Non-Negative Matrix Factorization (NMF) to extract the most relevant concepts used by the network (ResNet50V2) from the train set (ILSVRC2012~\cite{deng2009imagenet}). The global influence of these concepts on the predictions is then measured using Sobol indices (right panel). Finally, the method provides local explanations through \textit{concept attribution maps} (heatmaps associated with a concept, and computed using grad-CAM by backpropagating through the NMF concept values with implicit differentiation).
Besides, concepts can be interpreted by looking at crops that maximize the NMF coefficients. For the class ``chain saw'', the detected concepts seem to be:
  \textcolor{blue}{$\bullet$} the chainsaw engine, 
  \textcolor{purple}{$\bullet$} the saw blade, 
  \textcolor{green}{$\bullet$} the human head, 
  \textcolor{orange}{$\bullet$} the vegetation, 
  \textcolor{red}{$\bullet$} the jeans and
  \textcolor{dark}{$\bullet$} the tree trunk.
  \vspace{-4mm}
}
\label{fig:craft_demo}
\end{figure*}

Interpreting the decisions of modern machine learning models such as neural networks remains a major challenge. Given
the ever-increasing range of machine learning applications, the need for robust and reliable explainability methods continues to grow~\cite{doshivelez2017rigorous,jacovi2021formalizing}.
Recently enacted European laws (including the General Data Protection Regulation (GDPR)~\cite{kaminski2021right} and the European AI act~\cite{kop2021eu}) require the assessment of explainable decisions, especially those made by algorithms.

In order to try to meet this growing need, an array of explainability methods have already been proposed~\cite{saliency,GradCAM,smilkov2017smoothgrad,IntegratedGradients,shrikumar2017learning,deconvnet,guided-backprop,fel2021sobol,novello2022making}. One of the main class of methods called attribution methods yields heatmaps that indicate the importance of individual pixels for  driving a model's decision. 
However, these methods exhibit  critical limitations~\cite{sanity-checks,hase2021out,sixt2020explanations,fooling-lime}, as they have been shown to fail -- or only marginally help -- in recent human-centered benchmarks~\cite{fel2021cannot,kim2021hive,hase2020evaluating,nguyen2021effectiveness,shen2020useful,sixt2022users}. It has been suggested that their limitations stem from the fact that they are only capable of explaining \textit{where} in an image are the pixels that are critical to the decision but they cannot tell \textit{what} visual features are actually driving decisions at these locations. In other words, they show where the model looks but not what it sees. For example, in the scenario depicted in Fig.~\ref{fig:shovel}, where an ImageNet-trained ResNet mistakenly identifies an image as containing a shovel, the attribution map displayed on the left fails to explain the reasoning behind this misclassification.

A recent approach has sought to move past attribution methods~\cite{TCAV} by using so-called ``concepts'' to communicate information to users on how a model works. The goal is to find human-interpretable concepts in the activation space of a neural network. 
Although the approach exhibited potential, its practicality is significantly restricted due to the need for prior knowledge of pertinent concepts in its original formulation and, more critically, the requirement for a labeled dataset of such concepts.
Several lines of work have focused on trying to automate the concept discovery process based only on the training dataset and without explicit human supervision.
The most prominent of these techniques, ACE~\cite{ACE}, uses a combination of segmentation and clustering techniques but requires heuristics to remove outliers.
However, ACE provides  a proof of concept that it might be possible to discover concepts automatically and at scale -- without additional labeling or human supervision.
Nevertheless, the approach suffers several limitations: by construction, each image segment can only belong to a single cluster, a layer has to be selected by the user to be used to retrieve the relevant concepts, and the amount of information lost during the outlier rejection phase can be a cause of concern. More recently, Zhang et al.~\cite{voleurs-didee-nmf} proposes to leverage matrix decompositions on internal feature maps to discover concepts.

\begin{figure*}[t!]
\centering\includegraphics[width=1.0\textwidth]{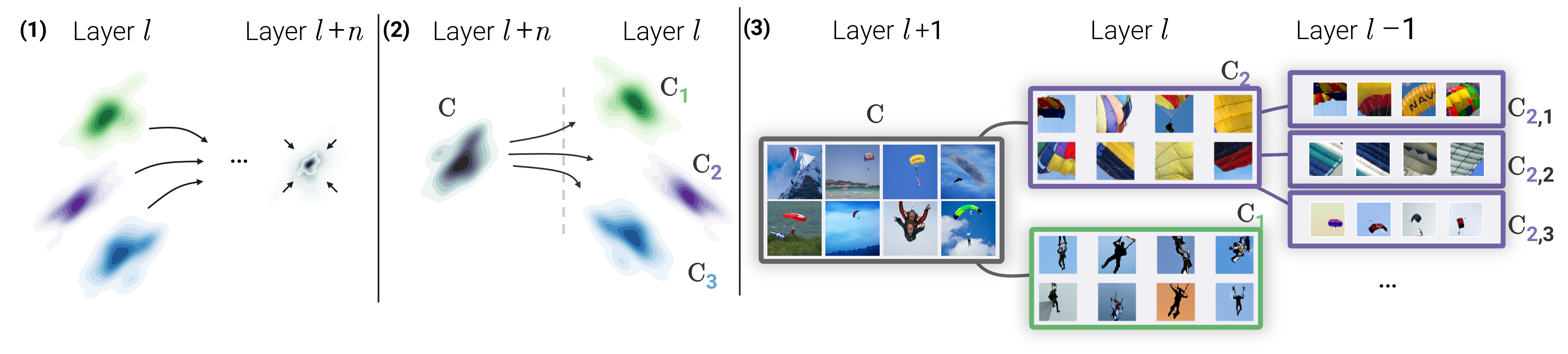}
\vspace{-1mm}
  \caption{ 
  \textbf{(1) Neural collapse (amalgamation).}
  A classifier needs to be able to linearly separate classes by the final layer. It is commonly assumed that in order to achieve this, image activations from the same class get progressively ``merged'' such that these image activations converge to a one-hot vector associated with the class at the level of the logits layer. 
  In practice, this means that different concepts get ultimately blended together along the way. 
  \textbf{(2) Recursive process.} When a concept is not understood (e.g., $\mathcal{C}$), we propose to decompose it into multiple sub-concepts (e.g., $\mathcal{C}_{\textcolor{green}{1}}, \mathcal{C}_{\textcolor{purple}{2}}, \mathcal{C}_{\textcolor{blue}{3}}$) using the activations from an earlier layer to overcome the aforementioned neural collapse issue.
  \textbf{(3) Example of recursive concept decomposition} using CRAFT on the ImageNet class ``parachute''.
  \vspace{-3.5mm}
  }
  \label{fig:collapse}
\end{figure*}

Here, we try to fill these gaps with a novel method called CRAFT which uses Non-Negative Matrix Factorization (NMF)~\cite{lee1999learning} for concept discovery. In contrast to other concept-based explanation methods, our approach provides an explicit link between their global and local explanations (Fig.~\ref{fig:craft_demo}) and identifies the relevant layer(s) to use to represent individual concepts (Fig.~\ref{fig:collapse}). Our main contributions can be described as follows:

{\textbf{(i)}} A novel approach for the automated extraction of high-level concepts learned by deep neural networks. We validate its practical utility to users with human psychophysics experiments.

{\textbf{(ii)}} A recursive procedure to automatically identify concepts and sub-concepts at the right level of granularity -- starting with our decomposition at the top of the model and working our way upstream. We validate the benefit of this approach with human psychophysics experiments showing that (i) the decomposition of a concept yields more coherent sub-concepts and (ii) that the groups of points formed by these sub-concepts are more refined and appear meaningful to humans.

{\textbf{(iii)}} A novel technique to quantify the importance of individual concepts for a model's prediction using Sobol indices~\cite{sobol1993sensitivity,sobol-review,sobol2001global} -- a technique borrowed from Sensitivity Analysis.

{\textbf{(iv)}} The first concept-based explainability method which produces  \textit{concept attribution maps} by backpropagating  concept scores  into the pixel space by leveraging the implicit function theorem in order to localize the pixels associated with the concept of a given input image. This effectively opens up the toolbox of both white-box~\cite{smilkov2017smoothgrad, saliency, IntegratedGradients, GradCAM, guided-backprop, CAM, fel2022eva} and black-box~\cite{LIME, unified-shapley, RISE, fel2021sobol} explainability methods to derive concept-wise attribution maps.

\section{Related Work}

\paragraph{Attribution methods}

Attribution methods are widely used as post-hoc explainability techniques to determine the input variables that contribute to a model's prediction by generating importance maps, such as the ones shown in Fig.\ref{fig:shovel}. The first attribution method, Saliency, introduced in~\cite{saliency}, generates a heatmap by utilizing the gradient of a given classification score with respect to the pixels. This method was later improved upon in the context of deep convolutional networks for classification in subsequent studies, such as~\cite{deconvnet, guided-backprop, IntegratedGradients, smilkov2017smoothgrad}.
However, the image gradient only reflects the model's operation within an infinitesimal neighborhood around an input, which can yield misleading importance estimates~\cite{ghalebikesabi2021locality} since gradients of large vision models are notoriously noisy~\cite{smilkov2017smoothgrad}. 
That is why several methods leverage perturbations on the input image to probe the model and create importance maps that indicate the most crucial areas for the decision, such as Rise~\cite{RISE}, Sobol~\cite{fel2021sobol}, or more recently HSIC~\cite{novello2022making}.

Unfortunately, a severe limitation of these approaches -- apart from the fact that they only show the ``\textit{where}'' -- is that they are subject to confirmation bias: while they may appear to offer useful explanations to a user, sometimes these explanations are actually incorrect~\cite{sanity-checks, ghorbani2017interpretation, fooling-lime}.
These limitations raise questions about their usefulness, as recent research has shown by using human-centered experiments to evaluate the utility of attribution~\cite{hase2020evaluating,nguyen2021effectiveness,fel2021cannot,kim2021hive,shen2020useful}.

In particular, in~\cite{fel2021cannot}, a protocol is proposed to measure the usefulness of explanations, corresponding to how much they help users identify rules driving a model's predictions (correct or incorrect) that transfer to unseen data -- using the concept of meta-predictor (also called simulatability)~\cite{kim2016examples,doshivelez2017rigorous,fong2017interpretable}. 
The main idea is to train users to predict the output of the system using a small set of images along with associated model predictions and corresponding explanations. 
A method that performs well on this this benchmark is said useful, as it help users better predict the output of the model by providing meaningful information about the internal functioning of the model.
This framework being agnostic to the type of explainability method, we have chosen to use it in Section~\ref{sec:exp} in order to compare CRAFT with attribution methods.

\vspace{-3mm}
\paragraph{Concepts-based methods}
Kim et al.~\cite{TCAV} introduced a method aimed at providing explanations that go beyond attribution-based approaches by measuring the impact of pre-selected concepts on a model's outputs. Although this method appears more interpretable to human users than standard attribution techniques, it requires a database of images describing the relevant concepts to be manually curated.
Ghorbani et al.~\cite{ACE} further extended the approach to extract concepts  without the need for human supervision. The approach, called ACE~\cite{ACE}, uses a segmentation scheme on images, that belong to an image class of interest. %
The authors leveraged the intermediate activations of a neural network for specific image segments. These segments were resized to the appropriate input size and filled with a baseline value. The resulting activations were clustered to produce prototypes, which they referred to as "concepts". However, some concepts contained background segments, leading to the inclusion of uninteresting and outlier concepts. To address this, the authors implemented a postprocessing cleanup step to remove these concepts, including those that were present in only one image of the class and were not representative. While this improved the interpretability of their explanations to human subjects, the use of a baseline value filled around the segments could introduce biases in the explanations~\cite{hsieh2020evaluations,sturmfels2020visualizing,haug2021baselines,kindermans2019reliability}.

Zhang et al.~\cite{voleurs-didee-nmf} developed a solution to the unsupervised concept discovery problem by using matrix factorizations in the latent spaces of neural networks. However, one major drawback of this method is that it operates at the level of convolutional kernels, leading to the discovery of localized concepts. For example, the concept of "grass" at the bottom of the image is considered distinct from the concept of "grass" at the top of the image.

\vspace{-2mm}
\section{Overview of the method} \label{sec:method}

In this section, we first describe our concept activations factorization method. Below we highlight the main differences with related work.
We then proceed to introduce the three novel ingredients that make up CRAFT: %
(1) a method to recursively decompose concepts into sub-concepts, 
(2) a method to better estimate the importance of extracted concepts, and 
(3) a method to use any attribution method to create \textit{concept attribution maps}, using implicit differentiation~\cite{krantz2002implicit,griewank2008evaluating,blondel2021implicitdiff}.%

\vspace{-3mm}

\paragraph{Notations}
In this work, we consider a general supervised learning setting, where $(\bm{x}_1, ..., \bm{x}_n) \in \mathcal{X}^n \subseteq \mathbb{R}^{n \times d}$ are $n$ inputs images and $(y_1, ..., y_n) \in \mathcal{Y}^n$ their associated labels. 
We are given a (machine-learnt) black-box predictor $\pred : \mathcal{X} \to \mathcal{Y}$, which at some test input $\bm{x}$ predicts the output $\pred(\bm{x})$.
Without loss of generality, we establish that $\pred$ is a neural network that can be decomposed into two distinct components. The first component is a function $\vg$ that maps the input to an intermediate state, and the second component is $\vh$, which takes this intermediate state to the output, such that $\pred(\bm{x}) = (\vh \circ \vg)(\bm{x})$. In this context, $\vg(\bm{x}) \subseteq \mathbb{R}^p$ represents the intermediate activations of $\bm{x}$ within the network.
Further, we will assume non-negative activations: $ \vg(\bm{x}) \geq 0$. In particular, this assumption is verified by any architecture that utilizes \textit{ReLU}, but any non-negative activation function works.

\subsection{Concept activation factorization.}\label{subsec:caf}

\begin{figure*}[t!]
\centering
\centering
  \includegraphics[width=1.0\textwidth]{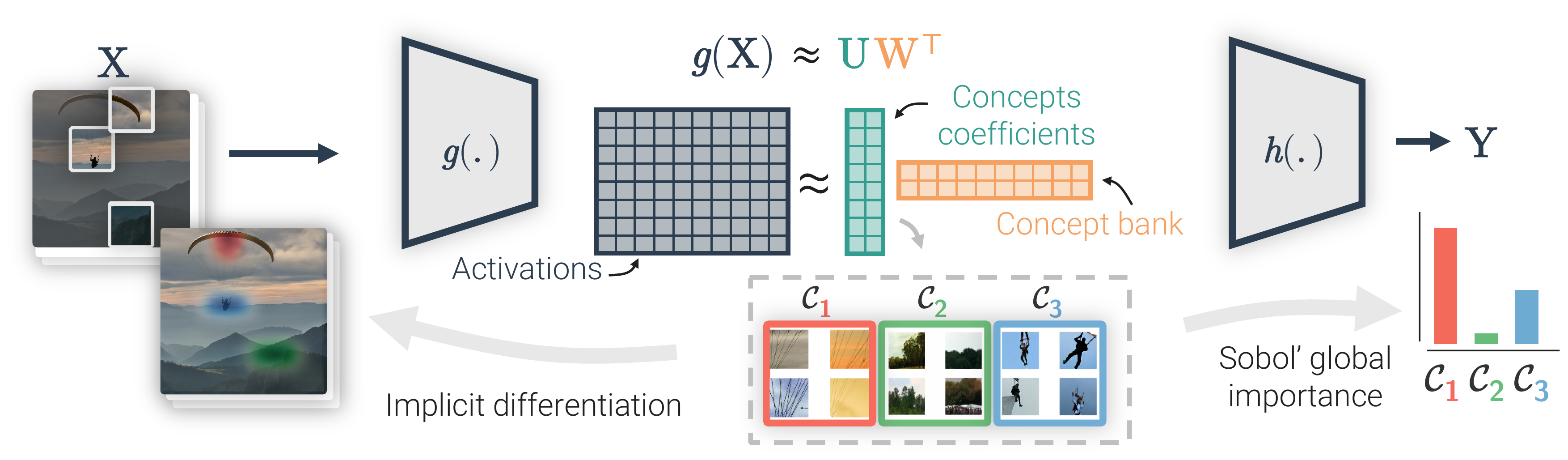}
  \vspace{-1mm}
  \caption{
  \textbf{Overview of CRAFT.}
  Starting from a set of crops $\vx$ containing a concept $\mathcal{C}$ (e.g., crops images of the class ``parachute''), we compute activations $\vg(\vx)$ corresponding to an intermediate layer from a neural network for random image crops. 
  We then factorize these activations into two lower-rank matrices, $(\textcolor{metalgreen}{\vu}, \textcolor{metalorange}{\vw})$. $\textcolor{metalorange}{\vw}$ is what we call a ``concept bank'' and is a new basis used to express the activations, while $\textcolor{metalgreen}{\vu}$ corresponds to the corresponding coefficients in this new basis.
  We then extend the method with 3 new ingredients: (1) recursivity -- by proposing to re-decompose a concept (e.g., take a new set of images containing $\mathcal{C}_{\textcolor{red}{1}}$) at an earlier layer, (2) a better importance estimation using Sobol indices and (3) an approach to leverage implicit differentiation to generate \textit{concept attribution maps} to localize concepts in an image.
  \vspace{-3.5mm}
  }
  \label{fig:craft}
\end{figure*}

We use Non-negative matrix factorization to identify a basis for concepts based on a network's activations (Fig.\ref{fig:craft}). Inspired by the approach taken in ACE~\cite{ACE}, we will use image sub-regions to try to identify coherent concepts. 

The first step involves gathering a set of images that one wishes to explain, such as the dataset, in order to generate associated concepts. In our examples, to explain a specific class $y \in \mathcal{Y}$, we selected the set of points $\mathcal{C}$ from the dataset for which the model's predictions matched a specific class $\mathcal{C} = \{ \bm{x}_i : \pred(\bm{x}_i) = y, 1 \leq i \leq n \}$.
It is important to emphasize that this choice is significant. The goal is not to understand how humans labeled the data, but rather to comprehend the model itself. By only selecting correctly classified images, important biases and failure cases may be missed, preventing a complete understanding of our model.

Now that we have defined our set of images, we will proceed with selecting sub-regions of those images to identify specific concepts within a localized context. It has been observed that the implementation of segmentation masks suggested in ACE can lead to the introduction of artifacts due to the associated inpainting with a baseline value.
In contrast, our proposed method takes advantage of the prevalent use of modern data augmentation techniques such as randaugment, mixup, and cutmix during the training of current models.
These techniques involve the current practice of models being trained on image crops, which enables us to leverage a straightforward crop and resize function denoted by $\bm{\pi}(\cdot)$ to create sub-regions (illustrated in Fig.\ref{fig:craft}). By applying $\bm{\pi}$ function to each image in the set $\mathcal{C}$, we obtain an auxiliary dataset $\vx \in \mathbb{R}^{n \times d}$ such that each entries $\vx_i = \bm{\pi}(\bm{x}_i)$ is an image crop.

To discover the concept basis, we start by obtaining the activations for the random crops $\va = \vg(\vx) \in \mathbb{R}^{n \times p}$.
In the case where $\pred$ is a convolutional neural network, a global average pooling is applied to the activations.

We are now ready to apply Non-negative Matrix Factorization (NMF) to decompose  positive activations $\va$ into a product of non-negative, low-rank matrices $\vu \in \mathbb{R}^{n \times r}$ and $\vw \in \mathbb{R}^{p \times r}$ by solving:
\vspace{-2.5mm}
\begin{equation}
\label{eq:nmf}
(\vu, \vw) = \argmin_{\vu \geq 0, \vw \geq 0} ~ \frac{1}{2}\|\va - \vu\vw^\mathsf{T} \|^2_{F}, %
\vspace{-2mm}
\end{equation}  
where $||\cdot||_F$ denotes the Frobenius norm.

This decomposition of our activations $\va$ yields two matrices: $\vw$ containing our Concept Activation Vectors (CAVs) and $\vu$ that redefines the data points in our dataset according to this new basis. Moreover, this decomposition in this new basis has some interesting properties that go beyond the simple low-rank factorization -- since $r \ll \min(n,p)$.
First, NMF can be understood as the joint learning of a dictionary of Concept Activation Vectors -- called a ``concept bank'' in Fig.~\ref{fig:craft} -- that maps a $\Reals^p$ basis onto $\Reals^r$, and $\vu$ the coefficients of the vectors $\va$ expressed in this new basis. 
The minimization of the reconstruction error $\frac{1}{2}\|\va - \vu\vw\|^2_F$ ensures that the new basis contains (mostly) relevant concepts. Intuitively, the non-negativity constraints $\vu \geq 0, \vw \geq 0$ encourage (\textbf{\textit{i}}) $\vw$ to be sparse (useful for creating disentangled concepts), (\textbf{\textit{ii}})  $\vu$ to be sparse (convenient for selecting a minimal set of useful concepts)  and (\textbf{\textit{iii}})  missing data to be imputed~\cite{ren2020using}, which corresponds to the sparsity pattern of \textit{post-ReLU} activations $\va$. 

It is worth noting that each input $\bm{x}_i$ can be expressed as a linear combination of concepts denoted as $\va_i = \sum_{j=1}^r \vu_{(i,j)} \vw_j^\mathsf{T}$. This approach is advantageous because it allows us to interpret each input as a composition of the underlying concepts. Furthermore, the strict positivity of each term -- NMF is working over the anti-negative semiring, -- enhances the interpretability of the decomposition. Another interesting interpretation could be that each input is represented as a superposition of concepts~\cite{elhage2022superposition}.

While other methods in the literature solve a similar problem (such as low-rank factorization using SVD or ICA), the NMF is both fast and effective and is known to yield concepts that are meaningful to humans~\cite{6165290,fu2019nonnegative,voleurs-didee-nmf}. Finally, once the concept bank $\vw$ has been precomputed, we can associate the concept coefficients $\bm{u}$ to any new input $\bm{x}$ (e.g., a full image) by solving the underlying Non-Negative Least Squares (NNLS) problem $\min_{\bm{u} \geq 0} ~ \frac{1}{2}\|\vg(\bm{x}) - \bm{u}\vw^\mathsf{T}\|^2_{F}$, and therefore recover its decomposition in the concept basis.

In essence, the core of our method can be summarized as follows: using a set of images, the idea is to re-interpret their embedding at a given layer as a composition of concepts that humans can easily understand. 
In the next section, we show how one can recursively apply concept activation factorizations to preceding layer for an image containing a previously computed concept.

\vspace{-1mm}
\subsection{Ingredient 1: A pinch of recursivity}\label{subsec:rec}

One of the most apparent issues in previous work~\cite{ACE,voleurs-didee-nmf} is the need for choosing a priori a layer at which the activation maps are computed. This choice will critically affect the concepts that are identified  because certain concepts get amalgamated~\cite{neural-collapse} into one at different layers of the neural network, resulting in incoherent and indecipherable clusters, as illustrated in Fig.~\ref{fig:collapse}. We posit that this can be solved by iteratively applying our decomposition at different layer depths, and for the concepts that remain difficult to understand, by looking for their sub-concepts in earlier layers by isolating the images that contain them. This allows us to build hierarchies of concepts for each class.

We offer a simple solution consisting of reapplying our method to a concept by performing a second step of concept activation factorization on a set of images that contain the concept $\mathcal{C}$ in order to refine it and create sub-concepts (e.g., decompose $\mathcal{C}$ into $\{ \mathcal{C}_1,\mathcal{C}_2,\mathcal{C}_3 \}$) see Fig.~\ref{fig:collapse} for an illustrative example. 
Note that we generalize current methods in the sense that taking images $(\bm{x}_1, ..., \bm{x}_n)$ that are clustered in the logits layer (belonging to the same class) and decomposing them in a previous layer -- as done in \cite{ACE, voleurs-didee-nmf} -- is a valid recursive step.
For a more general case, let us assume that a set of images that contain a common concept is obtained using the first step of concept activation factorization. 

We will then take a subset of the auxiliary dataset points to refine any concept $j$. To do this, we select the subset of points that contain the concept $\mathcal{C}_j = \{\vx_i : \vu_{(i,j)} > \lambda_j, 1 \leq i \leq n \}$, where $\lambda_j$ is the 90th percentile of the values of the concept $\vu_{(1,j)}, \ldots, \vu_{(n,j)}$. In other words, the 10\% of images that activate the concept $j$ the most are selected for further refinement into sub-concepts.
Given this new set of points, we can then re-apply the Concept Matrix Factorization method to an earlier layer $\vg'(\cdot)$ to obtain the sub-concepts decomposition from the initial concept -- as illustrated in Fig.\ref{fig:collapse}.

\vspace{-1mm}
\subsection{Ingredient 2: A dash of sensitivity analysis}\label{subsec:sobol}

A major concern with concept extraction methods is that concepts that makes sense to humans are not necessarily the same as those being used by a model to classify images.
In order to prevent such confirmation bias during our concept analysis phase, a faithful estimate the overall importance of the extracted concepts is crucial. 
Kim et al.~\cite{TCAV} proposed an importance estimator based on directional derivatives: the partial derivative of the model output with respect to the vector of concepts. 
While this measure is theoretically grounded, it relies on the same principle as gradient-based methods, and thus, suffers from the same pitfalls: neural network models have noisy gradients~\cite{smilkov2017smoothgrad,IntegratedGradients}. Hence, the farther the chosen layer is from the output, the noisier the directional derivative score will be.

Since we essentially want to know which concept has the greatest effect on the output of the model, it is natural to consider the field of sensitivity analysis~\cite{sobol-review, sobol1993sensitivity, sobol2001global, cukier1973study,idrissi2021developments}.
In this section, we briefly recall the classic ``total Sobol indices'' and how to apply them to our problem. The complete derivation of the Sobol-Hoeffding decomposition is presented in Section~\ref{apdx:sobol} of the supplementary materials.
Formally, a natural way to estimate the importance of a concept $i$ is to measure the fluctuations of the model's output $\vh(\vu \vw^\mathsf{T})$ in response to meaningful perturbations of the concept coefficient $\vu_{(1,i)}, \ldots, \vu_{(n,i)}$.
Concretely, we will use perturbation masks $\vm  = (M_1, ..., M_r) \sim \mathcal{U}([0, 1]^r)$, here an i.i.d sequence of real-valued random variables, we introduce a concept fluctuation to reconstruct a perturbed activation $\tilde{\va} = (\vu \odot \vm)\vw^\mathsf{T}$ where $\odot$ denote the Hadamard product (e.g., the masks can be used to remove a concept by setting its value to zero). We can then propagate this perturbed activation to the model output $\vy = \vh(\tilde{\va})$.
Simply put, removing or applying perturbation of an important concept will result in a substantial variation in the output, whereas an unused concept will have minimal effect on the output.

Finally, we can capture the importance that a concept might have as a main effect -- along with its interactions with other concepts -- on the model's output by calculating the expected variance that would remain if all the concepts except the $i$ were to be fixed. This yields the general definition of the total Sobol indices.

\vspace{-1mm}
\begin{definition}[\textbf{Total Sobol indices}]
    \textit{The total Sobol index $\mathcal{S}^T_i$, which measures the contribution of a concept $i$ as well as its interactions of any order with any other concepts to the model output variance, is given by:}
    \vspace{-1mm}
    \begin{align}
    \label{eq:total_sobol}
       \mathcal{S}^T_i 
      & = \frac{ \mathbb{E}_{\vm_{\sim i}}( \Var_{M_i} ( \vy | \vm_{\sim i} )) }{ \Var(\vy) } \\
       & = \frac{ \mathbb{E}_{\bm{M}_{\sim i}}( \Var_{M_i} ( \vh((\vu \odot \vm)\vw^\mathsf{T}) | \vm_{\sim i} )) }{ \Var( \vh((\vu \odot \vm)\vw^\mathsf{T})) }.
    \end{align}
\end{definition}
\vspace{-1mm}

In practice, this index can be calculated very efficiently~\cite{saltelli2010variance, marrel2009calculations, janon2014asymptotic, owen2013better, tarantola2006random}, more details on the Quasi-Monte Carlo sampling and the estimator used are left in appendix~\ref{apdx:sobol}.

\subsection{Ingredient 3: A smidgen of implicit differentiation}\label{subsec:cam}
\vspace{-2mm}
Attribution methods are useful for determining the regions deemed important by a model for its decision, but they lack information about what exactly triggered it.
We have seen that we can already extract this information from the matrices $\vu$ and $\vw$, but as it is, we do not know in what part of an image a given concept is represented.
In this section, we will show how we can leverage attribution methods (forward and backward modes) to find where a concept is located in the input image (see Fig.~\ref{fig:craft_demo}). Forward attribution methods do not rely on any gradient computation as they only use inference processes, whereas backward methods require back-propagating through a network's layers. By application of the chain rule, computing $\partial \vu / \partial \vx$ requires access to $\partial \vu /\partial \va$.

\vspace{-0.5mm}

To do so, one could be tempted to solve the linear system $\vu\vw^\mathsf{T}=\va$. 
However, this problem is ill-posed since $\vw^\mathsf{T}$ is low rank. A standard approach is to calculate the Moore-Penrose pseudo-inverse $(\vw^\mathsf{T})^{\dagger}$, which solves rank deficient systems by looking at the minimum norm solution~\cite{barata2012moore}. In practice, $(\vw^\mathsf{T})^{\dagger}$ is computed with the Singular Value Decomposition (SVD) of $\vw^\mathsf{T}$. Unfortunately, SVD is also the solution to the \textit{unstructured minimization} of $\frac{1}{2}\|\va-\vu\vw^\mathsf{T}\|^2_F$ by the Eckart-\-Young-\-Mirsky theorem~\cite{eckart1936approximation}. Hence, the non-negativity constraints of the NMF are ignored, which prevents such approaches from succeeding. Other issues stem from the fact that the $\vu,\vw$ decomposition is generally not unique.

Our third contribution consists of tackling this problem to allow the use of attribution methods, i.e., \textit{concept attribution maps}, by proposing a strategy to differentiate through the NMF block.

\vspace{-2mm}

\begin{figure*}[h!]
  \includegraphics[width=0.99\textwidth]{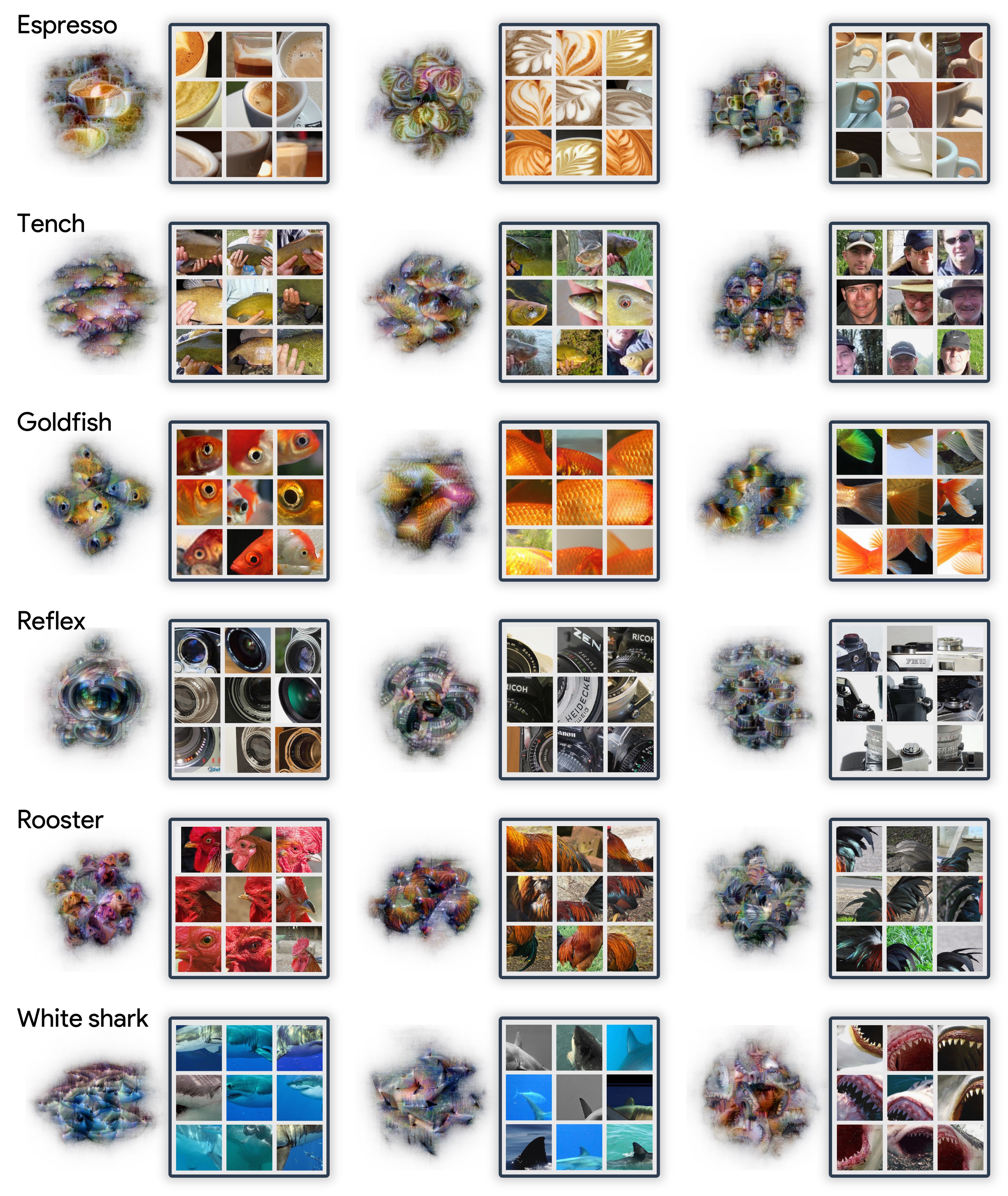}
  \caption{
  \textbf{Qualitative Results:} CRAFT results on 6 classes of ILSVRC2012~\cite{deng2009imagenet} for a trained ResNet50V2. The results showcase the top 3 most important concepts for each class. This is done by displaying crop images that activate the concept the most (using $\vu$) and also feature visualization~\cite{feature-viz} of the associated CAVs (using $\vw$). 
  }
  \label{fig:qualitative}
\end{figure*}

\paragraph{Implicit differentiation of NMF block}

The NMF problem~\ref{eq:nmf} is NP-hard~\cite{vavasis2010complexity}, and it is not convex with respect to the input pair $(\vu,\vw)$. However, fixing the value of one of the two factors and optimizing the other turns the NMF formulation into a pair of Non-Negative Least Squares (NNLS) problems, which are convex. This ensures that alternating minimization (a standard approach for NMF) of $(\vu,\vw)$ factors will eventually reach a local minimum.
Each of this alternating NNLS problems fulfills the Karush-–Kuhn-–Tucker (KKT) conditions~\cite{karush1939minima,kuhn1951nonlinear}, which can be encoded in the so-called \textit{optimality function} $\mathbf{F}$ from \cite{blondel2021implicitdiff}, see Eq.~\ref{apeq:optimality_fun} Appendix~\ref{app:implicit}. The implicit function theorem~\cite{griewank2008evaluating} allows us to use implicit differentiation~\cite{krantz2002implicit,griewank2008evaluating,bell2008algorithmic} to efficiently compute the Jacobians $\partial \vu/ \partial \va$ and $\partial \vw / \partial \va$ without requiring to back-propagate through each of the iterations of the NMF solver:
\vspace{-3mm}
\begin{equation}
    \frac{\partial (\vu,\vw,\bar \vu,\bar \vw)}{\partial \va}=-(\partial_1 \mathbf{F})^{-1}\partial_2 \mathbf{F}.
\end{equation}
\vspace{-2mm}

However, this requires the dual variables $\bar \vu$ and $\bar \vw$, which are not computed in scikit-learn's~\cite{pedregosa2011scikit} popular implementation\footnote{Scikit-learn uses a block coordinate descent algorithm~\cite{cichocki2009fast,fevotte2011algorithms}, with a randomized SVD initialization.}. Consequently, we leverage the work of~\cite{huang2016flexible} and we re-implement our own solver with Jaxopt~\cite{blondel2021implicitdiff} based on ADMM~\cite{boyd2011distributed}, a GPU friendly algorithm (see Appendix~\ref{app:implicit}).

\begin{table*}[t]
    \centering
    \resizebox{0.9\textwidth}{!}{%
        \begin{tabular}{c l cccc cccc cccc}
        \toprule
          & & \multicolumn{4}{c}{\textit{Husky vs. Wolf}} & \multicolumn{4}{c}{\textit{Leaves}} & \multicolumn{4}{c}{\textit{``Kit Fox'' vs ``Red Fox''}} \\
         \midrule
       & Session~n$^{\circ}$               & 1 & 2 & 3 & \metric & 1 & 2 & 3 & \metric & 1 & 2 & 3 & \metric \\
         \midrule
        & Baseline                    & 55.7 & 66.2 & 62.9 &  &       70.1 & 76.8 & 78.6 &  &       58.8 & 62.2 & 58.8 &  \\
        & Control                             & 53.3 & 61.0 & 61.4 & 0.95 &       72.0 & 78.0 & 80.2 & 1.02 &     60.7 & 59.2 & 48.5 & 0.94 \\
        \midrule
        \multirow{6}{*}{\rotatebox[origin=c]{90}{{\footnotesize Attributions}}}
        & Saliency~\cite{saliency}                  & 53.9 & 69.6 & 73.3 & 1.06  &        83.2 & 88.7 & 82.4 & \underline{1.13} &      61.7 & 60.2 & 58.2 & 1.00 \\ 
        & Integ.-Grad.~\cite{IntegratedGradients}     & 67.4 & 72.8 & 73.2 & 1.15 &      82.5 & 82.5 & 85.3 & 1.11 &       59.4 & 58.3 & 58.3 & 0.98\\
        & SmoothGrad~\cite{smilkov2017smoothgrad}           & 68.7 & 75.3 & 78.0 & 1.20 &    83.0 & 85.7 & 86.3 & \underline{1.13} &       50.3 & 55.0 & 61.4 & 0.93 \\
        & GradCAM~\cite{GradCAM}               & 77.6 & 85.7 & 84.1 & 1.34 &       81.9 & 83.5 & 82.4 & 1.10 &      54.4 & 52.5 & 54.1 & 0.90 \\
        & Occlusion~\cite{deconvnet}            & 71.0 & 75.7 & 78.1 & 1.22 &       78.8 & 86.1 & 82.9 & 1.10 &     51.0 & 60.2 & 55.1 & 0.92 \\
        & Grad.-Input~\cite{shrikumar2017learning}               & 65.8 & 63.3 & 67.9 & 1.06 &      76.5 & 82.9 & 79.5 & 1.05 &      50.0 & 57.6 & 62.6 & 0.95 \\

        \midrule
        
        \multirow{3}{*}{\rotatebox[origin=c]{90}{{\footnotesize Concepts}}}
        
        & ACE~\cite{ACE} & 68.8 & 71.4 & 72.7 & 1.15 &   79.8 & 73.8 & 82.1 & 1.05 &    48.4 & 46.5 & 46.1 & 0.78 \\

        & CRAFTCO (ours) & 82.4 & 87.0 & 85.1 & \underline{1.38} &  78.8 & 85.5 & 89.4 & 1.12 &    55.5 & 49.5 & 53.3 & 0.88 \\
        & CRAFT (ours) & 90.6 & 97.3 & 95.5 & \textbf{1.53} &   86.2 & 86.6 & 85.5 & \textbf{1.15} &    56.5 & 50.6 & 49.4 & 0.87 \\
        \bottomrule
        \end{tabular}%
    }
    \caption{\textbf{\metric~scores on 3 datasets from 
    \cite{fel2021cannot}}. Their \metric~benchmark evaluates how well explanations help users identify general rules driving classifications that readily transfer to unseen instances. At training time, users are asked to infer rules driving the decisions of the model given a set of images, and their associated predictions and explanations. At test time, the \metric~metric measures the accuracy of users at predicting the model decision on novel images averaged over 3 sessions, and normalized by the baseline accuracy of users trained without explanations.
    The higher the \metric~score, the more useful the explanation, and the more crucial the information provided is for understanding --and thus predicting the model's output-- on novel samples. %
    CRAFTCO stands for ``CRAFT Concept Only'' and designates an experimental condition where only global concepts are given to users, without local explanations (i.e., the concept attribution maps).
    The first and second best results above the baseline are in \textbf{bold} and \underline{underlined}, respectively. 
    }
    \label{tab:utility}
    \vspace{-3mm}
\end{table*}

Concretely, given our concepts bank $\vw$, the concept attribution maps of a new input $\bm{x}$ are calculated by solving the NNLS problem $\min_{\vu \geq 0} \frac{1}{2}\|\vg(\bm{x})-\vu\vw^\mathsf{T}\|^2_F$. The implicit differentiation of the NMF block $\partial \vu / \partial \va$ is integrated into the classic back-propagation to obtain $\partial \vu / \partial \bm{x}$. Most interestingly, this technical advance enables the use of all white-box explainability methods~\cite{smilkov2017smoothgrad, saliency, IntegratedGradients, GradCAM, guided-backprop, CAM} to generate concept-wise attribution maps and trace the part of an image that triggered the detection of the concept by the network. Additionally, it is even possible to employ black-box methods~\cite{LIME, RISE, unified-shapley, fel2021sobol} since it only amounts to solving an NNLS problem. %

\vspace{-1mm}
\section{Experimental evaluation}\label{sec:exp}

In order to evaluate the interest and the benefits brought by CRAFT, we start in Section~\ref{sec:expUtility} by assessing the practical utility of the method on a human-centered benchmark composed of 3 XAI scenarios.

After demonstrating the usefulness of the method using these human experiments, we independently validate the 3 proposed ingredients.
First, we provide evidence that recursivity allows refining concepts, making them more meaningful to humans using two additional human experiments in Section \ref{sec:expRecursivity}.
Next, we evaluate our new Sobol estimator and show quantitatively that it provides a more faithful assessment of concept importance in Section~\ref{sec:expSobol}.
Finally, we run an ablation experiment that measures the interest of local explanations based on concept attribution maps coupled with global explanations.
Additional experiments, including a sanity check and an example of deep dreams applied on the concept bank, as well as many other examples of local explanations for randomly picked images from ILSVRC2012, are included in Section~\ref{apx:more-craft} of the supplementary materials.
We leave the discussion on the limitations of this method and on the broader impact in appendix~\ref{apx:limitations}.

\subsection{Utility Evaluation}
\label{sec:expUtility}

As emphasized by Doshi-Velez et al.~\cite{doshivelez2017rigorous}, the goal of XAI should be to develop methods that help a user better understand the behavior of deep neural network models. An instantiation of this idea was recently introduced by Colin \& Fel et al.~\cite{fel2021cannot} who described an experimental framework to quantitatively measure the practical usefulness of explainability methods in real-world scenarios. For their initial setup, these authors recruited  $n=1,150$ online participants (evaluated over 8 unique conditions and 3 AI scenarios) -- making it the largest benchmark to date in XAI. Here, we follow their framework rigorously to allow for the robust evaluation of the utility of our proposed CRAFT method and the related ACE.
The 3 representative real-world scenarios are: (1) identifying bias in an AI system (using Husky vs Wolf dataset from~\cite{LIME}), (2) characterizing the visual strategy that are too difficult for an untrained non-expert human observer (using  the Paleobotanical dataset from \cite{wilf2016computer}), (3) understanding complex failure cases (using ImageNet ``Red fox'' vs ``Kit fox'' binary classification).
Using this benchmark, we evaluate CRAFT, ACE, as well as CRAFT with only the global concepts (CRAFTCO) to allow for a fair comparison with ACE.
To the best of our knowledge, we are the first to systematically evaluate concept-based methods against attribution methods.

Results are shown in Table~\ref{tab:utility} and demonstrate the benefit of CRAFT, which achieves higher scores than all of the attribution methods tested as well as ACE in the first two scenarios. To date, no method appears to exceed the baseline on the third scenario suggesting that additional work is required.
We also note that, in the first two scenarios, CRAFTCO is one of the best-performing methods and it always outperforms ACE -- meaning that even without the local explanation of the concept attribution maps, CRAFT largely outperforms ACE. Examples of concepts produced by CRAFT are shown in the Appendix~\ref{ap:utility}.

\subsection{Validation of Recursivity}
\label{sec:expRecursivity}

\begin{table}
\centering
    \resizebox{\linewidth}{!}{
    \begin{tabular}{lll}
    \toprule
    & Experts ($n=36$) & Laymen ($n=37$)\\
    \cmidrule[0.1pt](lr){2-3}
    \textit{Intruder}  \\
    \cmidrule[0.1pt](lr){1-3}
    Acc. Concept     & 70.19\%  & 61.08\%     \\
    Acc. Sub-Concept & 74.81\% ($p = 0.18$)  & \textbf{67.03}\% ($p = 0.043$)      \\
    \midrule
    \textit{Binary choice} \\
    \cmidrule[0.1pt](lr){1-3}
    Sub-Concept & \textbf{76.1}\% ($p < 0.001$) & \textbf{74.95}\% ($p < 0.001$)\\
    Odds Ratios & $3.53$ & $2.99$\\
    \bottomrule
    \end{tabular}
    }
\caption{\textbf{Results from the psychophysics experiments to validate the recursivity ingredient. \vspace{-5mm}}}\label{tab:results}
\end{table}

To evaluate the meaningfulness of the extracted high-level concepts, we performed psychophysics experiments with human subjects, whom we asked to answer a survey in two phases. Furthermore, we distinguished two different audiences: on the one hand, experts in machine learning, and on the other hand, people with no particular knowledge of computer vision. Both groups of participants were volunteers and did not receive any monetary compensation. Some examples of the developed interface are available the appendix~\ref{app:human-exp}. It is important to note that this experiment was carried out independently from the utility evaluation and thus it was setup differently.
\newline\textbf{Intruder detection experiment} First, we ask users to identify the intruder out of a series of five image crops belonging to a certain class, with the odd one being taken from a different concept but still from the same class. Then, we compare the results of this intruder detection with another intruder detection, this time, using a concept (e.g., $\mathcal{C}_1$) coming from a layer $l$ and one of its sub-concepts (e.g., $\mathcal{C}_{12}$ in Fig.\ref{fig:collapse}) extracted using our recursive method. If the concept (or sub-concept) is coherent, then it should be easy for the users to find the intruder.
Table~\ref{tab:results} summarizes our results, showing that indeed both concepts and sub-concepts are coherent, and that recursivity can lead to a slightly higher understanding of the generated concepts (significant for non-experts, but not for experts) and might suggest a way to make concepts more interpretable.
\newline\textbf{Binary choice experiment} In order to test the improvement of coherence of the sub-concept generated by recursivity with respect to the larger parent concept, we showed participants an image crop belonging to both a subcluster and a parent cluster (e.g.,  $\bm{\pi}(\bm{x}) \in \mathcal{C}_{11} \subset \mathcal{C}_1$) and asked them which of the two clusters (i.e., $\mathcal{C}_{11}$ or $\mathcal{C}_{1}$) seemed to accommodate the image the best. If our hypothesis is correct, then the concept refinement brought by recursivity should help form more coherent clusters.
The results in Table~\ref{tab:results} are satisfying since in both the expert and non-expert groups, the participants chose the sub-cluster more than 74\% of the time. We measure the significance of our results by fitting a binomial logistic regression to our data, and we find that both groups are more likely to choose the sub-concept cluster (at a $p < 0.001$).

\subsection{Fidelity analysis} \label{sec:expSobol}

\begin{figure}[h]
  \includegraphics[width=\linewidth]{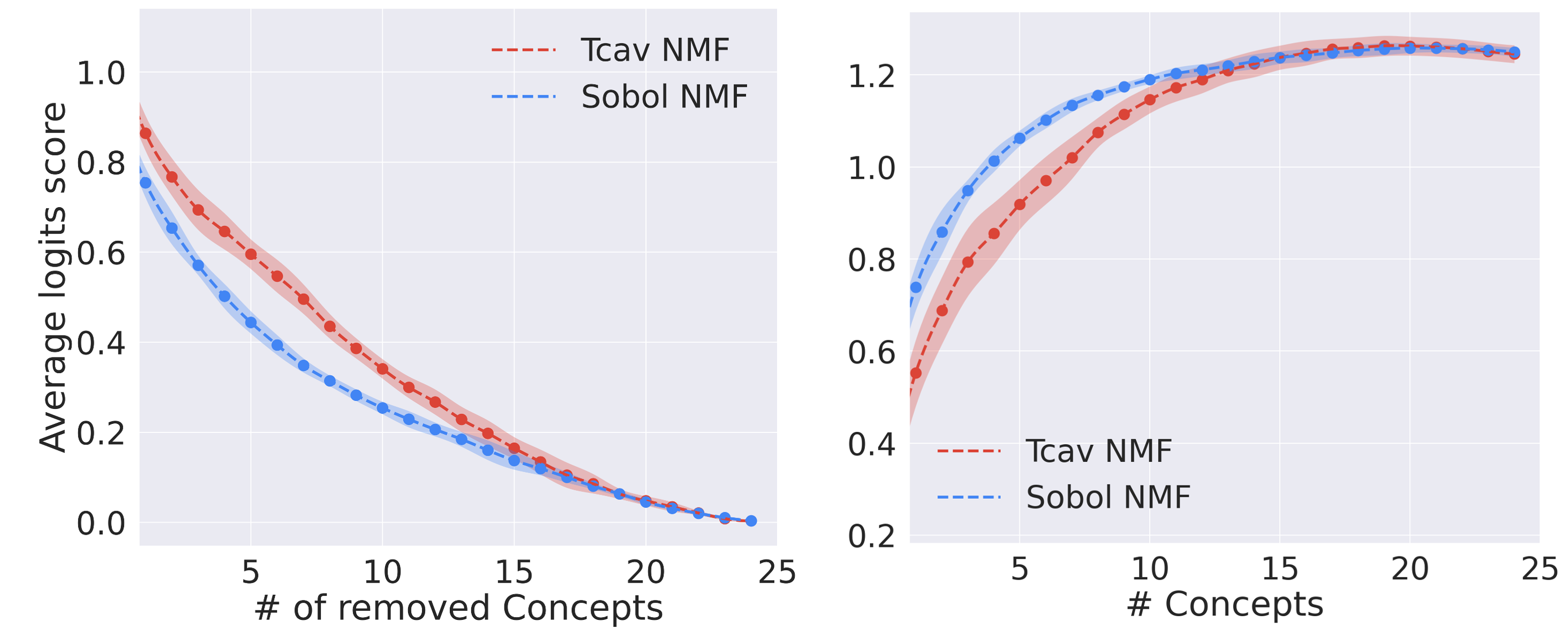}
  \caption{
  \textbf{(Left)} Deletion curves (lower is better). \textbf{(Right)} Insertion curves (higher is better). 
  For both the deletion or insertion metrics, Sobol indices lead to better estimates (calculated on >100K images) of important concepts. %
  \vspace{-4mm}
  }
  \label{fig:deletion}
\end{figure}

We propose to simultaneously verify that identified concepts are faithful to the model and that the concept importance estimator performs better than that used in TCAV~\cite{TCAV} by using the fidelity metrics introduced in \cite{ACE, voleurs-didee-nmf}. These metrics are similar to the ones used for attribution methods, which consist of studying the change of the logit score when removing/adding pixels considered important. Here, we do not introduce these perturbations in the pixel space but in the concept space: once $\vu$ and $\vw$ are computed, we reconstruct the matrix $\va\approx \vu\vw^\mathsf{T}$ using only the most important concept (or removing the most important concept for deletion) and compute the resulting change in the output of the model.  As can be seen from Fig.~\ref{fig:deletion}%
, ranking the extracted concepts using Sobol's importance score results in steeper curves than when they are sorted by their TCAV scores. %
We confirm that these results generalize with other matrix factorization techniques (PCA, ICA, RCA) in Section~\ref{app:fidelity} of the Appendix.

\vspace{-3mm}

\section{Conclusion}

In this paper, we introduced CRAFT, a method for automatically extracting human-interpretable concepts from deep networks. Our method aims to explain a pre-trained model's decisions both on a per-class and per-image basis by highlighting both ``what'' the model saw and ``where'' it saw it  -- with complementary benefits. The approach relies on 3 novel ingredients: 1) a recursive formulation of concept discovery to identify the correct level of granularity for which individual concepts are understandable; 2) a novel method for measuring concept importance through Sobol indices to more accurately identify which concepts influence a model's decision for a given class; and 3) the use of implicit differentiation methods to backpropagate through non-negative matrix factorization (NMF) blocks to allow the generation of concept-wise local explanations or \textit{concept attribution maps} independently of the attribution method used. Using a recently introduced human-centered utility benchmark, we conducted psychophysics experiments to confirm the validity of the approach: and that the concepts identified by CRAFT are useful and meaningful to human experimenters. We hope that this work will guide further efforts in the search for concept-based explainability methods.

\section*{Acknowledgements}

This work was conducted as part of the DEEL project.\footnote{\url{www.deel.ai}} Funding was provided by ANR-3IA Artificial and Natural Intelligence Toulouse Institute (ANR-19-PI3A-0004). Additional support provided by ONR grant N00014-19-1-2029 and NSF grant IIS-1912280. Support for computing hardware provided by Google via the TensorFlow Research Cloud (TFRC) program and by the Center for Computation and Visualization (CCV) at Brown University (NIH grant S10OD025181).

\clearpage

\newpage
\balance
{\small
\bibliographystyle{ieee_fullname}
\bibliography{bibliography.bib}
}

\newpage
\appendix

\clearpage
\setcounter{figure}{0}
\setcounter{table}{0}
\renewcommand\thefigure{S\arabic{figure}}
\renewcommand\thetable{S\arabic{table}}

\startcontents[annexes]
\printcontents[annexes]{l}{0}{\setcounter{tocdepth}{3}}
\clearpage

\section{Limitations and broader impact}\label{apx:limitations}

\subsection{Limitations}

Although we believe concept-based XAI to be a promising research direction, it isn't without pitfalls. It is capable of producing explanations that are ideally easy to understand by humans, but to what extent is a question that remains unanswered. The fact that there is no way to mathematically measure this prevents researchers from easily comparing the different techniques in the literature other than through time consuming and expensive experiments with human subjects. We think that developing a metric should be one of the field's priorities.%

With CRAFT, we address the question of \textit{what} by showing a cluster of the images that better represent each concept. However, we recognize that it's not perfect: in some cases, concepts are difficult to clearly define -- put a label on what it represents --, and might induce some confirmation and selection bias. Feature visualization~\cite{feature-viz} might help in better illustrating the specific concept (as done in appendix \ref{app:feature-viz-val}), but we believe there's still space for improvement. For instance, an interesting idea could be to leverage image captioning methods to describe the clusters of image crops, as textual information could help humans in better understanding clusters.

Although we believe CRAFT to be a considerable step in the good direction for the field of concept-based XAI, it also have some pitfalls. Namely, we chose the NMF as the activation factorization, which, while drastically improving the quality of extracted concepts, also comes with it's own caveats. For instance, it is known to be NP-hard to compute exactly, and in order to make it scalable, we had to use a tractable approximation by alternating the optimization of $\vu$ and $\vw$ through ADMM~\cite{boyd2011distributed}. This approach might indeed yield non-unique solutions. Our experiments (section \ref{sec:expSobol}), have shown a low variance on between the runs, which comforts us about the stability of our results.%
However the absence of formal guarantee for uniqueness must be kept in mind: this subject is still an active topic of research and improvement could be expected in the near future. Namely, sparsity constraints and regularization seem to be promising paths.
Naturally, we also need enough samples of the class under study to be available for the factorization to construct a relevant concept bank, which might affect the quality of the explanations on frugal applications where data is very scarce. %

\subsection{Broader impact}

We do hope that CRAFT helps in the transition to more human-understandable ways of explaining neural network models. It's capacity to find easily understandable concepts inside complex architectures and providing an indication of \textit{where} they are located in the image is -- to the best of our knowledge -- unmatched. 
We also think that this method's structure is a step towards reducing confirmation bias: for instance dataset's labels are never used in this method, only the model's predictions. Without claiming to remove confirmation bias, the method focuses on what \textit{the model sees} rather than what \textit{we expect the model to see}.
We believe this can help end-users build trust on computer vision models%
, and at the same time, provide ML practitioners with insights into potential sources of bias in the dataset (e.g. the ski pants in the astronaut/shovel example). Other methods in the literature obtaining similar results require very specific architectures~\cite{ProtoPNet} or to train another model to generate the explanations~\cite{ge2021peek}, so CRAFT provides a considerable advantage in the matter of flexibility in comparison.

\section{More results of CRAFT}\label{apx:more-craft}

\subsection{Qualitative comparison with ACE}\label{apx:qualitative}

Figure~\ref{fig:qualitative} compares the examples of concepts found by CRAFT against those found by ACE~\cite{ACE} for 3 classes of Imagenette. 
For each class the concepts are ordered by importance (the highest being the most important). 
ACE uses a clustering technique and TCAV to estimate importance, while CRAFT uses the method introduced in \ref{sec:method} and Sobol to estimate importance. These examples illustrate one of the weaknesses of ACE: the segmentation used can introduce biases through the baseline value used~\cite{sturmfels2020visualizing,meaningful-perturbations}. The concepts found by CRAFT seem distinct: (vault, cross, stained glass) for the Church class, (dumpster, truck door, two-wheeler) for the garbage truck, and (eyes, nose, fluffy ears) for the English Springer.

\begin{figure*}[hbtp]
  \includegraphics[width=0.99\textwidth]{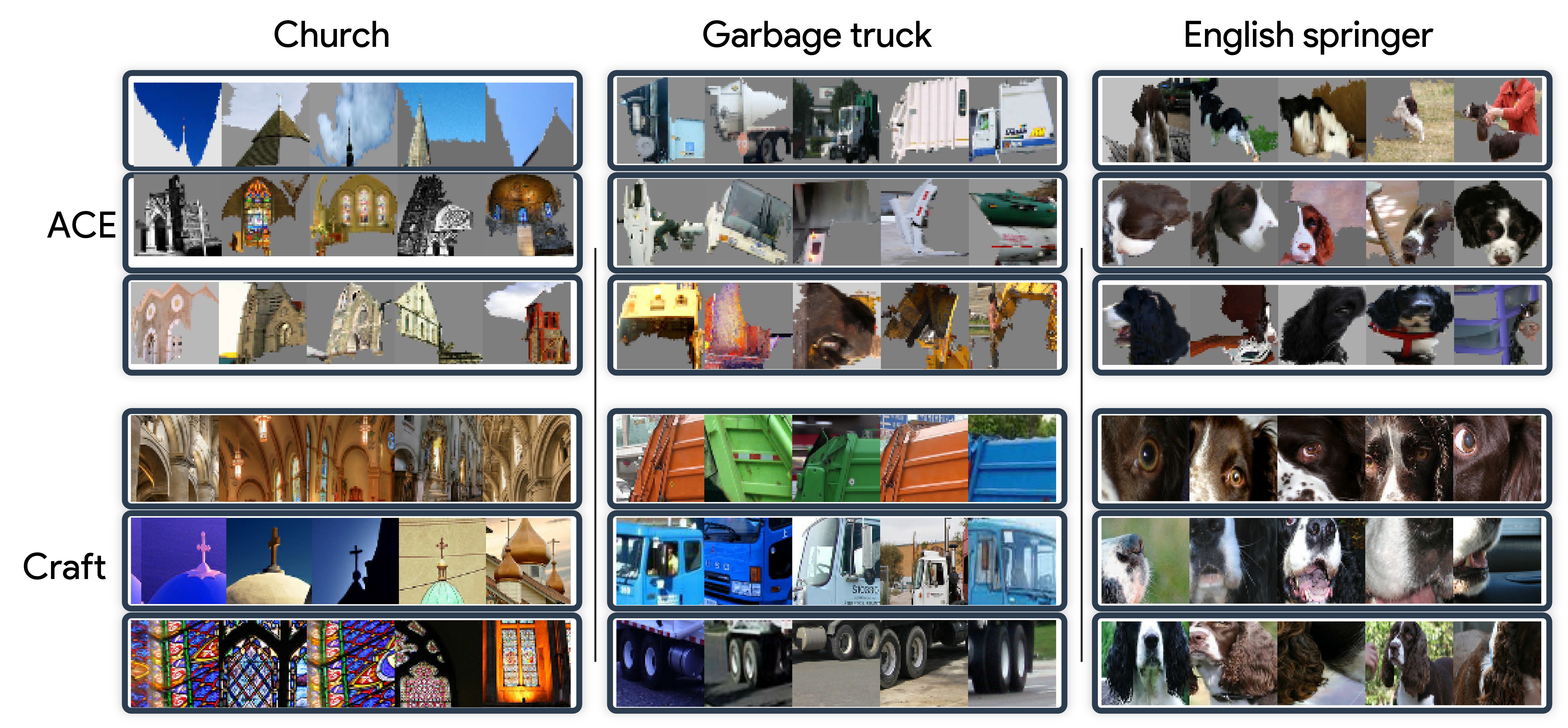}
  \caption{ \textbf{Qualitative comparison.} We compare concepts found by our method (top) to those extracted with ACE~\cite{ACE} (bottom) for the classes \textit{Church}, \textit{Garbage truck} and \textit{English springer} from ILSVRC2012~\cite{deng2009imagenet}. %
  }
  \label{fig:qualitative}
\end{figure*}

\subsection{Most important concepts.} 
\begin{figure*}[hbtp]
    \centering
      \includegraphics[width=\textwidth]{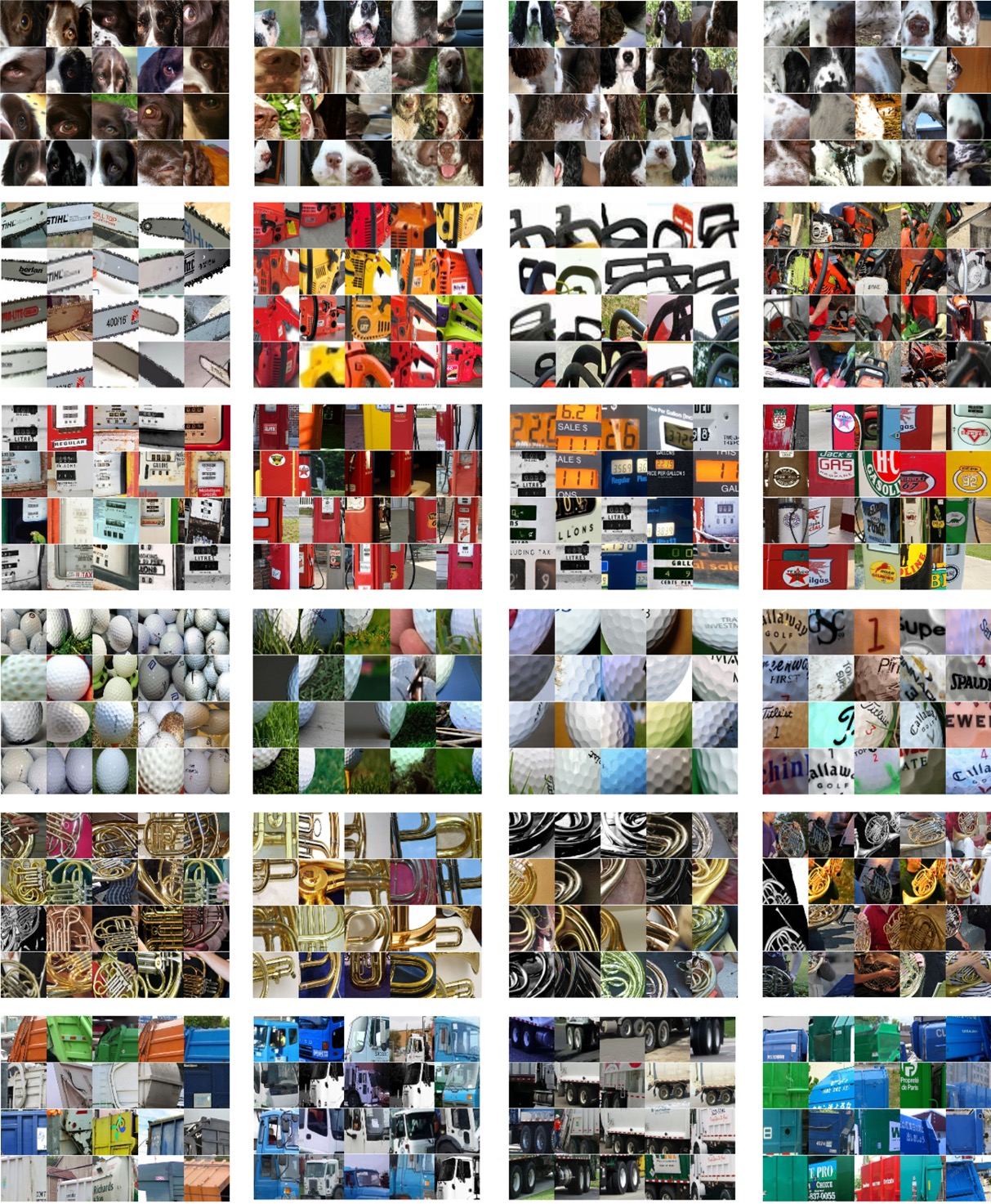}
      \caption{ \textbf{CRAFT most important concepts}. The 4 most important concepts ranked by importance (left to right) for the following classes: `English springer', `Chain saw',  `Gas pump', `Golf ball', `French horn',  and `Garbage truck'.
      }
      \label{fig:segments_all}
\end{figure*}

We show more example of the 4 most importants concepts for 6 classes: `Chain saw', `English springer', `Gas pump', `Golf ball', `French horn' and `Garbage Truck' (Figure~\ref{fig:segments_all}).%

\clearpage
\newpage

\subsection{Feature Visualization validation} \label{app:feature-viz-val}

\begin{figure}[H]
\centering
  \includegraphics[width=0.45\textwidth]{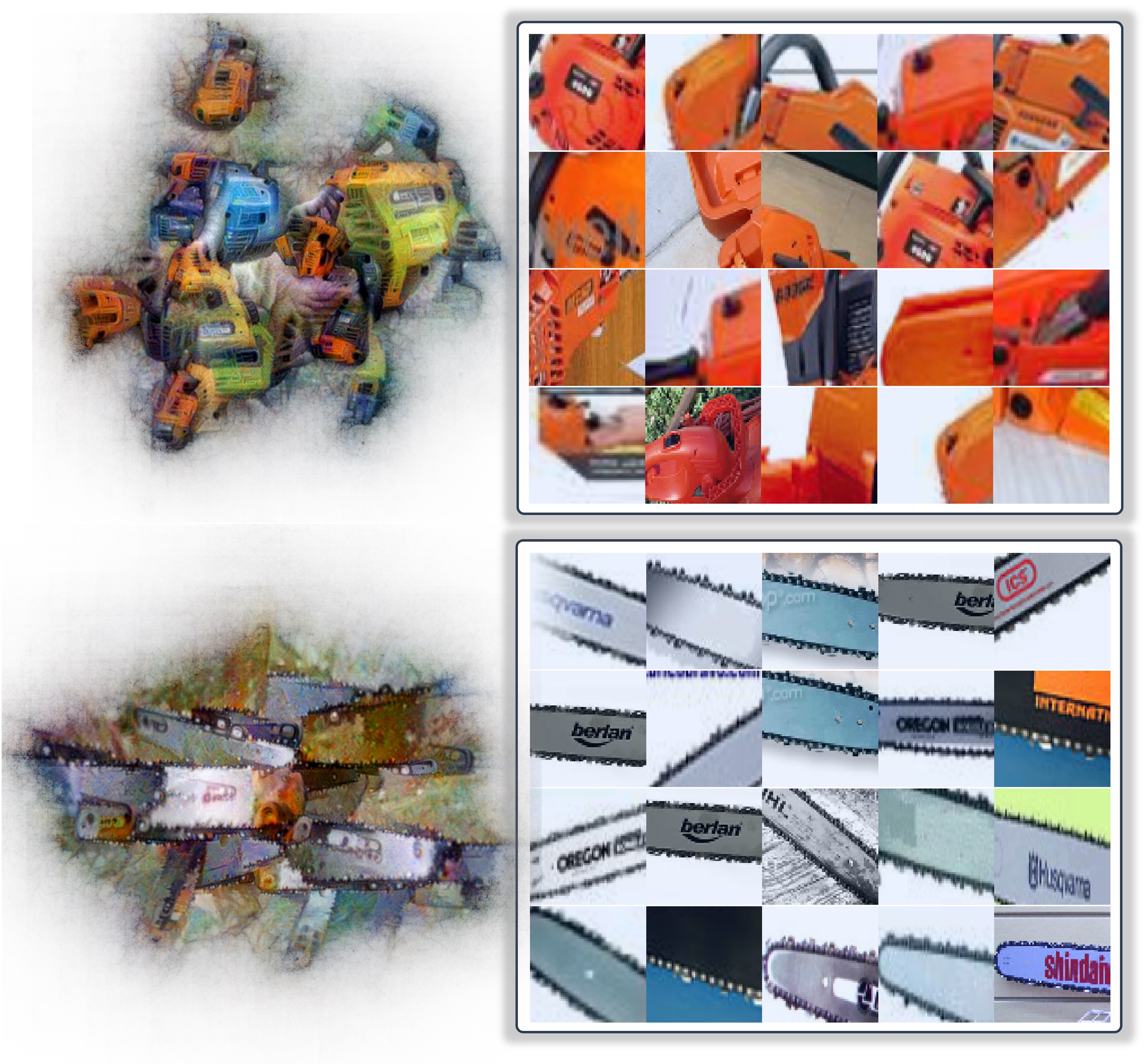}
  \caption{ \textbf{Feature visualization for chainsaw CRAFT concepts.}
  }
  \label{fig:feature_viz_chainsaw}
\end{figure}

\begin{figure}[H]
\centering
  \includegraphics[width=0.45\textwidth]{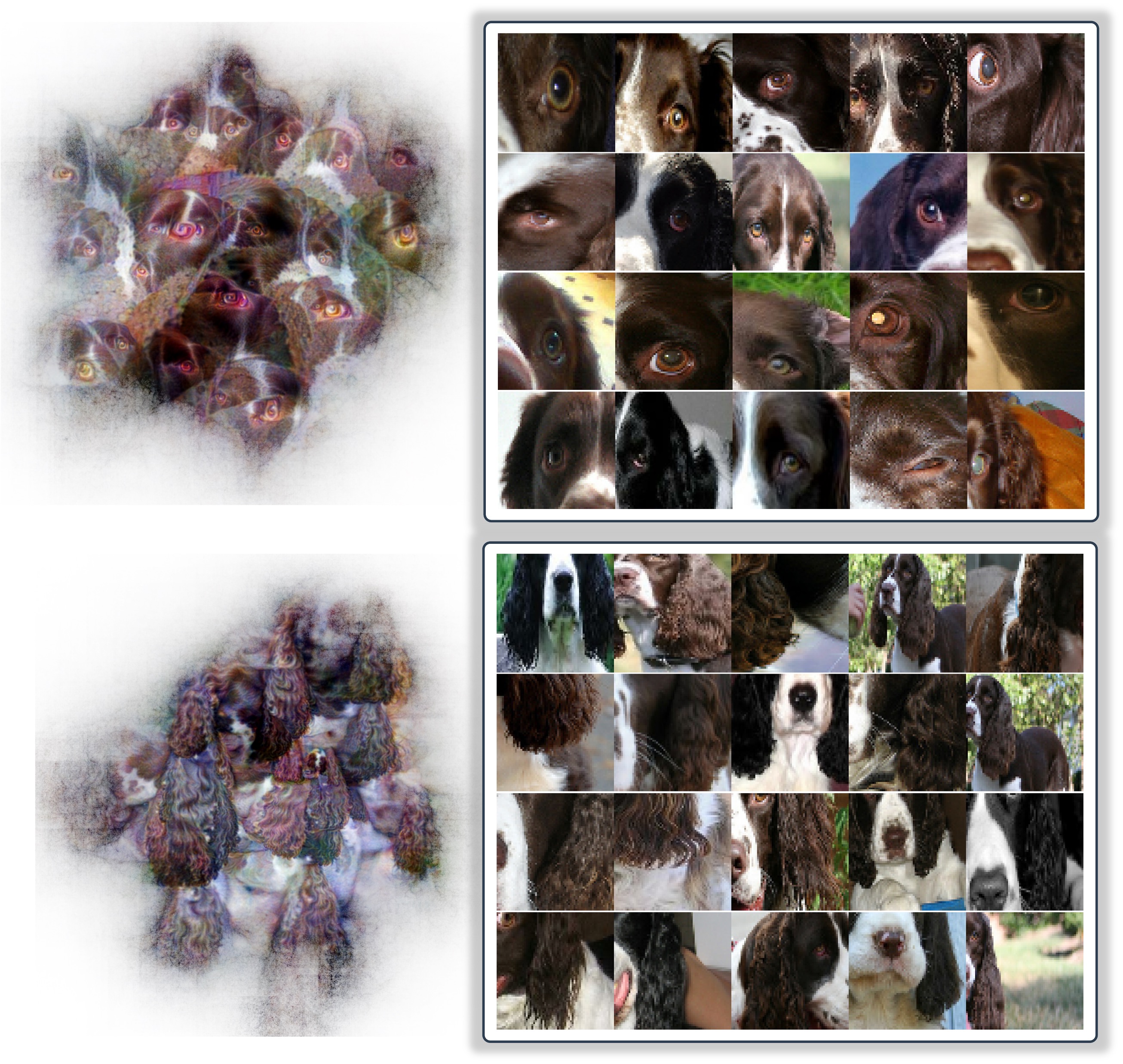}
  \caption{ \textbf{Feature visualization for english springer CRAFT concepts.}
  }
  \label{fig:feature_viz_englishspringer}
\end{figure}

\begin{figure}[H]
\centering
  \includegraphics[width=0.45\textwidth]{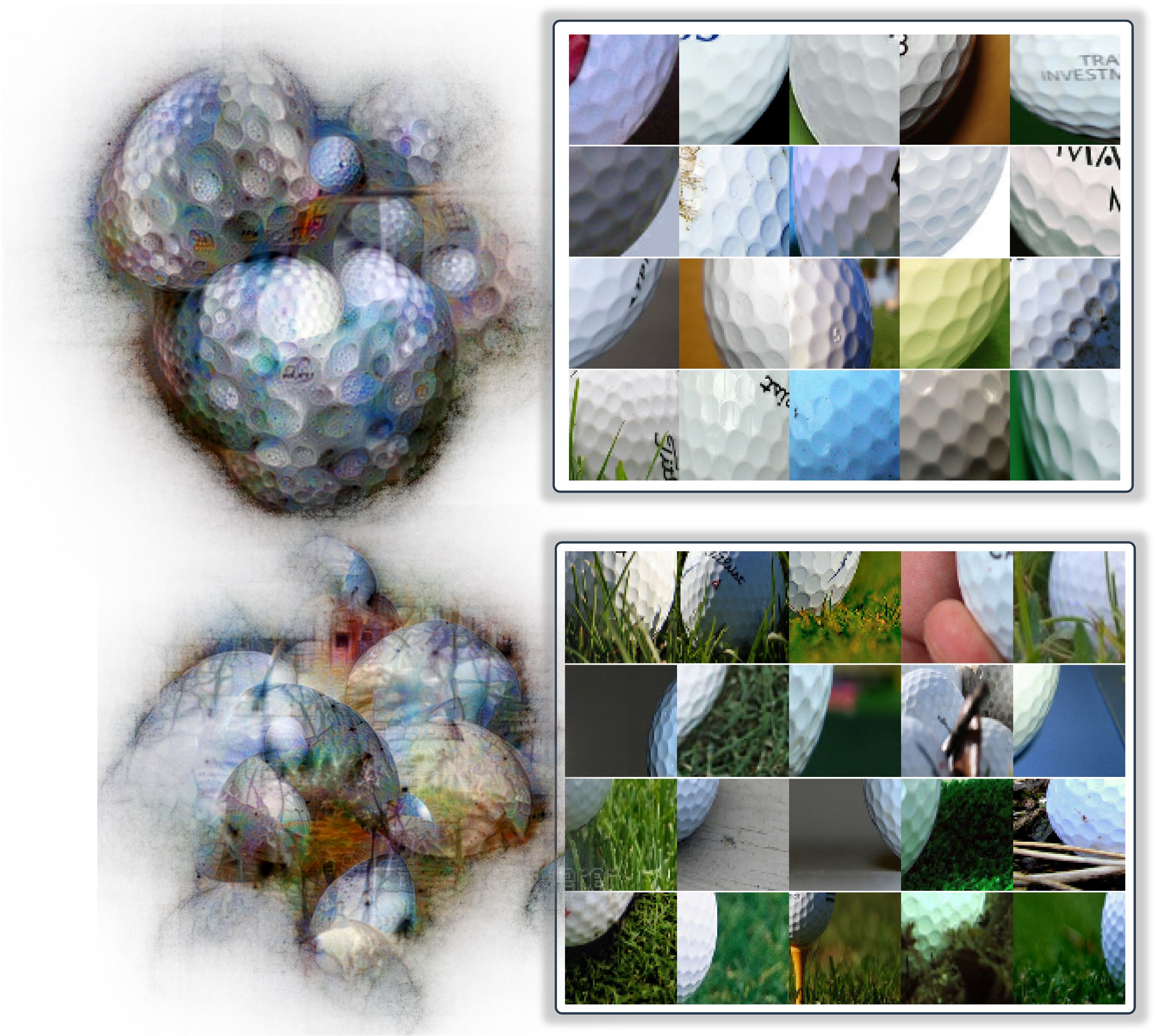}
  \caption{ \textbf{Feature visualization for golf CRAFT concepts.} 
  }
  \label{fig:feature_viz_golf}
\end{figure}

Another way of interpreting concepts -- as per~\cite{TCAV} -- is to employ feature visualization methods: through optimization, find an image that maximizes an activation pattern.
In our case, we used the set of regularization and constraints proposed by \cite{feature-viz}, which allow us to successfully obtain realistic images. In Figures~[\ref{fig:feature_viz_chainsaw}-\ref{fig:feature_viz_golf}], we showcase these synthetic images obtained through feature visualization, along with the segments that maximize the target concept. We observe that they do reflect the underlying concepts of interest.

Concretely, to produce those feature visualization, we are looking for an image $\bm{x}^*$ that is optimized to correspond to a concept from the concept bank $\vw_i$. We use the so called `dot-cossim' loss proposed by ~\cite{feature-viz}, which give the following objective:

\begin{equation*}
    \bm{x}^* = \argmax_{\bm{x} \in \mathcal{X}} ~ \langle \bm{g}(\bm{x}), \vw_i \rangle 
\frac{\langle \bm{g}(\bm{x}), \vw_i \rangle^2}{||\bm{g}(\bm{x})|| ~ ||\vw_i||  } - \mathcal{R}(\bm{x})    
\end{equation*}

With $\mathcal{R}(\cdot)$, the regularizations applied to $\bm{x}$  -- the default regularizations in the \textbf{Xplique} library~\cite{xplique}. As for the specific parameters, we used Fourier preconditioning on the image with a decay rate of $0.8$ and an Adam optimizer ($lr = 1e-1$). 

here

\clearpage
\newpage

\section{Backpropagating through the NMF block}

\subsection{Alternating Direction Method of Multipliers (ADMM) for NMF}

We recall that NMF decomposes the positive features vector $\va \in \mathbb{R}^{n \times p}$ of $n$ examples lying in dimension $p$, into a product of positive low rank matrices $\vu(\va)\in\mathbb{R}^{n\times r}$ and $\vw(\va)\in\mathbb{R}^{p\times r}$ (with $r<<\min(n,p)$), i.e the solution to the problem:
\begin{align}\label{apeq:nmf}
\min_{\vu\geq 0,\vw\geq 0} & \frac{1}{2}\|\va-\vu\vw^T\|^2_F. %
\end{align}

For simplicity we used a non-regularized version of the NMF objective, following Algorithms 1 and 3 in paper~\cite{huang2016flexible}, based on ADMM~\cite{boyd2011distributed}. This algorithm transforms the non-linear equality constraints into indicator functions $\bm{\delta}$. Auxiliary variables $\tilde \vu,\tilde \vw$ are also introduced to separate the optimization of the objective on the one side, and the satisfaction of the constraint on $\vu, \vw$ on the other side. The equality constraints $\tilde \vu=\vu,\tilde \vw=\vw$ are linear and easily handled by the ADMM framework through the associated dual variables $\bar \vu,\bar \vw$. In our case, the problem in Equation~\ref{apeq:nmf} is transformed into:
  
\begin{equation}
\begin{aligned}\label{apeq:admm}
\min_{\vu,\tilde \vu, \vw,\tilde \vw} & \frac{1}{2}\|\va-\tilde \vu \tilde \vw^T\|^2_F+\bm{\delta}(\vu)+\bm{\delta}(\vw), 
\\
~ ~ s.t. ~ ~ &\tilde \vu=\vu, \tilde \vw=\vw \\
     \text{with} ~ & \bm{\delta}(\bm{H})=\begin{cases}
                            0 \text{ if } \bm{H} \geq 0,\\
                            +\infty \text{ otherwise.}
                            \end{cases}
\end{aligned}
\end{equation}

Note that $\tilde \vu$ and $\vu$ (resp. $\tilde \vw$ and $\vw$) seem redundant: they are meant to be equal thanks to constraints $\tilde \vu=\vu, \tilde \vw=\vw$. This is standard practice within ADMM framework: introducing redundancies allows to disentangle the (unconstrained) optimization of the objective on one side (with $\tilde \vu$ and $\tilde \vw$) and constraint satisfaction on the other side with $\vu$ and $\vw$. During the optimization process the variables $\tilde \vu,\vu$ (resp. $\tilde \vw,\vw$) are different, and only become equal in the limit at convergence. The dual variables $\bar \vu,\bar \vw$ control the balance between optimization of the objective $\frac{1}{2}\|\va-\tilde \vu \tilde \vw^T\|^2_F$ and constraint satisfaction $\tilde \vu=\vu, \tilde \vw=\vw$. The constraints are simplified at the cost of a non-smooth (and even a non-finite) objective function $\frac{1}{2}\|\va -\bar \vu \bar \vw^T\|^2_F+\bm{\delta}(\vu)+\bm{\delta}(\vw)$ due to the term $\bm{\delta}(\vu)+\bm{\delta}(\vw)$. ADMM proceeds to create a so-called \textit{augmented Lagrangian} with $l_2$ regularization $\rho>0$:
\begin{equation}
    \begin{aligned}
    \Lagrangian&(\va,\vu,\vw,\tilde \vu,\tilde \vw,\bar \vu,\bar \vw)=\\
    &\frac{1}{2}\|\va-\tilde \vu\tilde \vw^T\|^2_F+\bm{\delta}(\vu)+\bm{\delta}(\vw)\\
    &+\bar \vu^T(\tilde \vu-\vu)+\bar \vw^T(\tilde \vw-\vw)\\
    &+\frac{\rho}{2}\left(\|\tilde \vu-\vu\|_2^2+\|\tilde \vw-\vw\|_2^2\right).
    \end{aligned}
\end{equation}

This regularization ensures that the dual problem is well posed and that it remain convex, even with the non smooth and infinite terms $\bm{\delta}(\vu)+\bm{\delta}(\vw)$. Once again, this is standard practice within ADMM framework. The (regularized) problem associated to this Lagrangian is decomposed into a sequence of convex problems that alternate minimization over the $\vu,\tilde \vu,\bar \vu$ and the $\vw,\tilde \vw,\bar \vw$ triplets.
  
\begin{align}\label{apeq:pairnnls}
\vu_{t+1}&=\argmin_{\vu=\tilde \vu} \frac{1}{2}\|\va-\tilde \vu\vw_t^T\|^2_F+\bm{\delta}(\vu)+\frac{\rho}{2}\|\tilde \vu-\vu\|_2^2. %
\\
\vw_{t+1}&=\argmin_{\vw=\tilde \vw} \frac{1}{2}\|\va-\vu_t\tilde \vw^T\|^2_F+\bm{\delta}(\vw)+\frac{\rho}{2}\|\tilde \vw-\vw\|_2^2.%
\end{align}

This guarantees a monotonic decrease of the objective function $\|\va-\tilde \vu_t\tilde \vw_t^T\|_F^2$. Each of these sub-problems is thus solved with ADMM separately, by alternating minimization steps of $\frac{1}{2}\|\va-\tilde \vu\vw_t^T\|^2_F+\bar \vu^T(\tilde \vu-\vu)+\frac{\rho}{2}\|\vu-\tilde \vu\|_2^2$ over $\tilde \vu$ (\textbf{\textit{i}}), with minimization steps of $\bm{\delta}(\vu)+\frac{\rho}{2}\|\vu-\tilde \vu\|_2^2$ over $\vu$ (\textbf{\textit{ii}}), and gradient ascent steps (\textbf{\textit{iii}}) on the dual variable $\bar \vu\gets \bar \vu+(\tilde \vu-\vu)$. A similar scheme is used for $\vw$ updates. Step (\textbf{\textit{i}}) is a simple convex quadratic program with equality constraints, whose KKT~\cite{karush1939minima,kuhn1951nonlinear} conditions yield a linear system with a Positive Semi-Definite (PSD) matrix. Step (\textbf{\textit{ii}}) is a simple projection of $\tilde \vu$ onto the convex set $\bm{\delta}^{-1}(\bm{0})$. Finally, step (\textbf{\textit{iii}}) is inexpensive.

Concretely, we solved the quadratic program using Conjugate Gradient~\cite{hestenes1952methods}, from \textit{jax.scipy.sparse.linalg.cg}. This indirect method only involves \textit{matrix-vector} products and can be more GPU-efficient than methods that are based on matrix factorization (such as Cholesky decomposition). Also, we re-implemented the pseudo code of~\cite{huang2016flexible} in \textit{Jax} for a fully GPU-compatible program. We used the primal variables $\vu_0,\vw_0$ returned by \textit{sklearn.decompose.nmf} as a \textit{warm start} for ADMM and observe that the high quality initialization of these primal variables considerably speeds up the convergence of the dual variables.

\subsection{Implicit differentiation}\label{app:implicit}

The Lagrangian of the NMF problem reads $\mathcal{L}(\vu,\vw,\bar \vu,\bar \vw)=\frac{1}{2}\|\va-\vu\vw^T\|_F^2-\bar \vu^T\vu-\bar \vw^T\vw$, with dual variables $\bar \vu$ and $\bar \vw$ associated to the constraints $\vu\geq 0, \vw \geq 0$. It yields a function $\bm{F}$ based on the KKT conditions~\cite{karush1939minima,kuhn1951nonlinear} whose optimal tuple $\vu,\vw,\bar \vu,\bar \vw$ is a root.  
  
For single NNLS problem (for example, with optimization over $\vu$) the KKT conditions are:

\begin{equation} %
    \begin{cases}
    \nabla_{\vu}\left(\frac{1}{2}\|\va-\tilde \vu \tilde \vw^T\|^2_F+\bar \vu^T(-\vu)\right)    =0, \text{ stationarity,}\\
    -\vu\leq 0, \text{ primal feasability,}\\
    \bar \vu \odot \vu   =0, \text{ complementary slackness,}\\
    \bar \vu   \geq 0, \text{ dual feasability.}\\
\end{cases}
\label{apeq:optimality_fun}
\end{equation}

By stacking the KKT conditions of the NNLS problems the we obtain the so-called \textit{optimality function} $\bm{F}$:

\begin{equation} %
    \bm{F}((\vu,\vw,\bar \vu,\bar \vw),\va)=\begin{cases}
    (\vu\vw^T-\va)\vw-\bar \vu    ,& \\ %
    (\vw\vu^T-\va^T)\vu-\bar \vw  ,& \\ %
    \bar \vu \odot \vu   ,& \\ %
    \bar \vw \odot \vw   .& \\ %
\end{cases}
\label{eq:optimality_fun_2}
\end{equation}

The implicit function theorem~\cite{griewank2008evaluating} allows us to use implicit differentiation~\cite{krantz2002implicit,griewank2008evaluating,bell2008algorithmic} to efficiently compute the Jacobians $\frac{\partial \vu}{\partial \va}$ and $\frac{\partial \vw}{\partial \va}$ without requiring to back-propagate through each of the iterations of the NMF solver:
\begin{equation}
    \frac{\partial (\vu,\vw,\bar \vu,\bar \vw)}{\partial \va}=-(\partial_1 \bm{F})^{-1}\partial_2 \bm{F}.
\end{equation}

Implicit differentiation requires access to the dual variables of the optimization problem in equation~\ref{eq:nmf}, which are not computed by Scikit-learn's popular implementation. Scikit-learn uses Block coordinate descent algorithm~\cite{cichocki2009fast,fevotte2011algorithms}, with a randomized SVD initialization. Consequently, we leverage our implementation in Jax based on ADMM~\cite{boyd2011distributed}.

Concretely, we perform a two-stage backpropagation \textit{Jax (2)}$\veryshortarrow$\textit{Tensorflow (1)} to leverage the advantage of each framework. The lower stage (1) corresponds to feature extraction $\va=\vh_l(\vx)$ from crops of images $\vx$, and upper stage (2) computes NMF $\va \approx \vu\vw^T$.  
  
We use the \textit{Jaxopt}~\cite{blondel2021implicitdiff} library that allows efficient computation of $\frac{\partial (\vu,\vw,\bar \vu,\bar \vw)}{\partial \va}=-(\partial_1 \bm{F})^{-1}\partial_2 \bm{F}$. The matrix $(\partial_1 \bm{F})^{-1}$ is never explicitly computed -- that would be too costly. Instead, the system $\partial_1 \bm{F}\frac{\partial (\vu,\vw,\bar \vu,\bar \vw)}{\partial \va}=-\partial_2 \bm{F}$ is solved with Conjugate Gradient~\cite{hestenes1952methods} through the use of Jacobian Vector Products (JVP) $\bm{v}\mapsto (\partial_1 \bm{F})\bm{v}$.  
  
The chain rule yields:
$$\frac{\partial \vu}{\partial \vx}=\frac{\partial \va}{\partial \vx}\frac{\partial \vu}{\partial \va}.$$

Usually, most Autodiff frameworks (e.g Tensorflow, Pytorch, Jax) handle it automatically. Unfortunately, combining two of those framework raises a new difficulty since they are not compatible. Hence, we re-implement manually the two stages auto-differentiation.  
  
Since $r$ is far smaller ($r=25$ in all our experiments) than input dimension $\vx$ (typically $224\times 244$ for ImageNet images), back-propagation is the preferred algorithm in this setting over forward-propagation. We start by computing sequentially the gradients $\nabla_{\vx} \vu_i$ for all concepts $1\leq i\leq r$. This amounts to compute $\bm{v}=\nabla_{\va} \vu_i$ with Implicit Differentiation in Jax, convert the Jax array $\bm{v}$ into Tensorflow tensor, and then to compute $\nabla_{\vx} \vu_i=\frac{\partial \va}{\partial \vx}\nabla_{\va} \vu_i=\nabla_{\vx} (\vh_l(\vx) \cdot \bm{v})$. The latter is easily done in Tensorflow. Finally we stack the gradients $\nabla_{\vx} \vu_i$ to obtain the Jacobian $\frac{\partial \vu}{\partial \vx}$.

\section{Sobol indices for concepts} \label{apdx:sobol}

We propose to formally derive the Sobol indices for the estimation of the importance of concepts.
Let us define a probability space  $(\Omega, \mathcal{F}, \mathbb{P})$ of possible concept perturbations. In order to build these concept perturbations, we start from an original vector of concepts coefficient $\widehat \vu \in \mathbb{R}^r$ and use i.i.d. stochastic masks $\vm = (M_1, ..., M_r) \sim \mathcal{U}([0, 1]^r)$, as well as a perturbation operator $\bm{\tau}(\cdot)$ to create stochastic perturbation of $\widehat \vu$ that we call concept perturbation $\vu = \bm{\tau}(\widehat \vu, \vm)$.

Concretely, to create our concept perturbation we consider the inpainting function as our perturbation operator (as in \cite{LIME, RISE, fel2021sobol}) : $\bm{\tau}(\vu, \vm) = \vu \odot \vm + (\bm{1} - \vm) \mu$ with $\odot$ the Hadamard product and $\mu \in \mathbb{R}$ a baseline value, here zero.
For the sake of notation, we will note $\pred : \mathcal{F} \to \mathbb{R}$ the function mapping a random concept perturbation $\vu$ from an intermediat layer to the output.
We denote the set $\mathcal{U} = \{1, ..., r\}$, $\bm{u}$ a subset of $\mathcal{U}$, its complementary $\sim \bm{u}$ and $\mathbb{E}(\cdot)$ the expectation over the perturbation space.
Finally, we assume that $\pred \in \mathbb{L}^2(\mathcal{F}, \mathbb{P})$ i.e. $|\mathbb{E}(\pred(\vu))| < + \infty$.

The Hoeffding decomposition allows us to express the function $\pred$ into summands of increasing dimension, denoting $\pred_{\bm{u}}$ the partial contribution of the concepts $\vu_{\bm{u}} = (U_i)_{i\in \bm{u}}$ to the score $\pred(\vu)$:
\begin{equation}
    \label{eq:anova}
    \begin{aligned}
    \bm{f}(\bm{U}) &= \bm{f}_{\emptyset}\\
    & + \sum_i^r \bm{f}_i(U_i)\\
    & + \sum_{1 \leqslant i < j \leqslant r} \bm{f}_{i,j}(U_i, U_j)
    & + \cdots \\
    & + \bm{f}_{1,...,r}(U_1, ..., U_r) \\
    &= \sum_{\substack{\bm{u} \subseteq \mathcal{U}}} \bm{f}_{\bm{u}}(\bm{U}_{\bm{u}}).
    \end{aligned}
\end{equation}

Eq.~\ref{eq:anova} consists of $2^r$ terms and is unique under the following orthogonality constraint:
\begin{equation}
    \label{eq:anova_ortho}
    \begin{aligned}
    \forall (\bm{u},\bm{v}) \subseteq \mathcal{U}^2 \; s.t. \;  \bm{u} \neq \bm{v}, \;\; \mathbb{E}\big(\bm{f}_{\bm{u}}(\bm{U}_{\bm{u}}) \bm{f}_{\bm{v}}(\bm{U}_{\bm{v}})\big) = 0.
    \end{aligned}
\end{equation}

Furthermore, orthogonality yields the characterization $\bm{f}_{\bm{u}}(\bm{U}_{\bm{u}}) = \mathbb{E}(\bm{f}(\bm{U})|\bm{U}_{\bm{u}}) - \sum_{\bm{v}\subset \bm{u}}\bm{f}_{\bm{v}}(\bm{U}_{\bm{v}})$ and allows us to decompose the model variance as:
\begin{equation}
    \label{eq:var_decomposition}
    \begin{aligned}
        \Var(\pred(\bm{U})) &= \sum_i^r \Var(\bm{f}_i(U_i)) \\
        &+\sum_{1 \leqslant i < j \leqslant r} \Var(\bm{f}_{i,j}(U_i, U_j))\\
        &+ ... + \Var(\bm{f}_{1,...,r}(U_1, ..., U_r)) \\
        &=\sum_{\substack{\bm{u} \subseteq \mathcal{U}}} \Var(\bm{f}_{\bm{u}}(\bm{U}_{\bm{u}})).
        \end{aligned}
\end{equation}

Building from Eq.~\ref{eq:var_decomposition}, it is natural to characterize the influence of any subset of concepts $\bm{u}$ as its own variance w.r.t. the total variance. This yields, after normalization by $\Var(\bm{f}(\bm{U}))$, the general definition of Sobol' indices.
\begin{definition}[Sobol indices~\cite{sobol1993sensitivity}]
\label{def:sobol_indice}
The sensitivity index $\mathcal{S}_{\bm{u}}$ which measures the contribution of the concept set $\bm{U}_{\bm{u}}$ to the model response $\bm{f}(\bm{U})$ in terms of fluctuation is given by:
\begin{equation}\label{eq:sobol_indice}
\begin{aligned}
    \mathcal{S}_{\bm{u}}  &= \frac{ \Var(\bm{f}_{\bm{u}}(\bm{U}_{\bm{u}})) }{ \Var(\pred(\bm{U})) }\\
    &= \frac{ \Var(\mathbb{E}(\bm{f}(\bm{U}) | \bm{U}_{\bm{u}})) - \sum_{\bm{v}\subset \bm{u}}\Var(\mathbb{E}(\bm{f}(\bm{U}) | \bm{U}_{\bm{v}} ))}{ \Var(\bm{f}(\bm{U})) }.
\end{aligned}
\end{equation}
\end{definition}

Sobol indices give a quantification of the importance of any subset of concepts with respect to the model decision, in the form of a normalized measure of the model output deviation from $\bm{f}(\bm{U})$. Thus, Sobol indices sum to one : $\sum_{\bm{u} \subseteq \mathcal{U}} \mathcal{S}_{\bm{u}} = 1$. 

\vspace{2mm}
Furthermore, the framework of Sobol' indices enables us to easily capture higher-order interactions between features. Thus, we can view the Total Sobol indices defined in \ref{eq:total_sobol} as the sum of of all the Sobol indices containing the concept $i$ : $\mathcal{S}^{T}_i = \sum_{\bm{u} \subseteq \mathcal{U}, i \in \bm{u}} \mathcal{S}_{\bm{u}}$. Concretely, we estimate the total Sobol indices using the Jansen estimator~\cite{janon2014asymptotic} and Quasi-Monte carlo Sequence (Sobol $LP_{\tau}$ sequence).

\clearpage

\section{Human experiments}\label{app:human-exp}

We first describe how participants were enrolled in our studies, then the general experimental design they went through.

\subsection{Utility evaluation}
\label{ap:utility}

\paragraph{Participants}
The participants that went through our experiments are users from the online platform Amazon Mechanical Turk (AMT), specifically, we recruit users with high qualifications (number of HIT completed $=5 000$ and HIT accepted $> 98 \%$). All participants provided informed consent electronically in order to perform the experiment ($\sim 5 - 8$ min), for which they received 1.4\$.\\

For the \textit{Husky vs. Wolf} scenario, $n=84$ participants passed all our screening and filtering process, respectively $n=32$ for CRAFT, $n=22$ for ACE and $n=22$ for CRAFTCO.

For the \textit{Leaves} scenario, after filtering, we analyzed data from $n=87$ participants, respectively $n=32$ for CRAFT, $n=24$ for ACE and $n=31$ for CRAFTCO.

For the \textit{"Kit Fox" vs. "Red Fox"} scenario, the results come from $n=79$ participants who passed all our screening processes, respectively $n=22$ for CRAFT, $n=31$ for ACE and $n=26$ for CRAFTCO.

\paragraph{General study design}
We followed the experimental design proposed by Colin and Fel et al.\cite{fel2021cannot}, in which explanations are evaluated according to their ability to help training participants at getting better at predicting their models' decisions on unseen images.

Each of those participants are only tested on a single condition to avoid possible experimental confounds. 

The main experiment is divided into 3 training sessions (with 5 training samples in each) each followed by a brief test. In each individual training trial, an image was presented with the associated prediction of the model, together with an explanation. After a brief training phase (5 samples), participants' ability to predict the classifier's output was evaluated on 7 new samples during a test phase. During the test phase, no explanation was provided.
We also use the reservoir that subjects can refer to during the testing phase to minimize memory load as a confounding factor.

We implement the same 3-stage screening process as in \cite{fel2021cannot}: First we filter participants not successful at the practice session done prior to the main experiment used to teach them the task, then we have them go through a quiz to make sure they understood the instructions. Finally, we add a catch trial in each testing phase --that users paying attention are expected to be correct on-- allowing us to catch uncooperative participants.

\begin{figure*}[hb]
    \centering
    \begin{subfigure}{0.9\textwidth}
        \centering
        \includegraphics[width=0.45\textwidth]{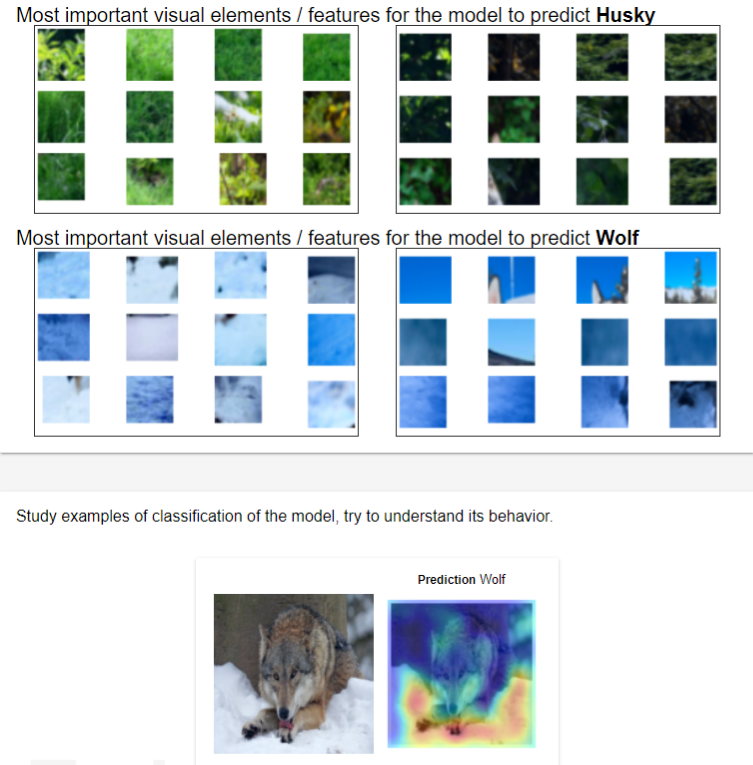}
        \includegraphics[width=0.45\textwidth]{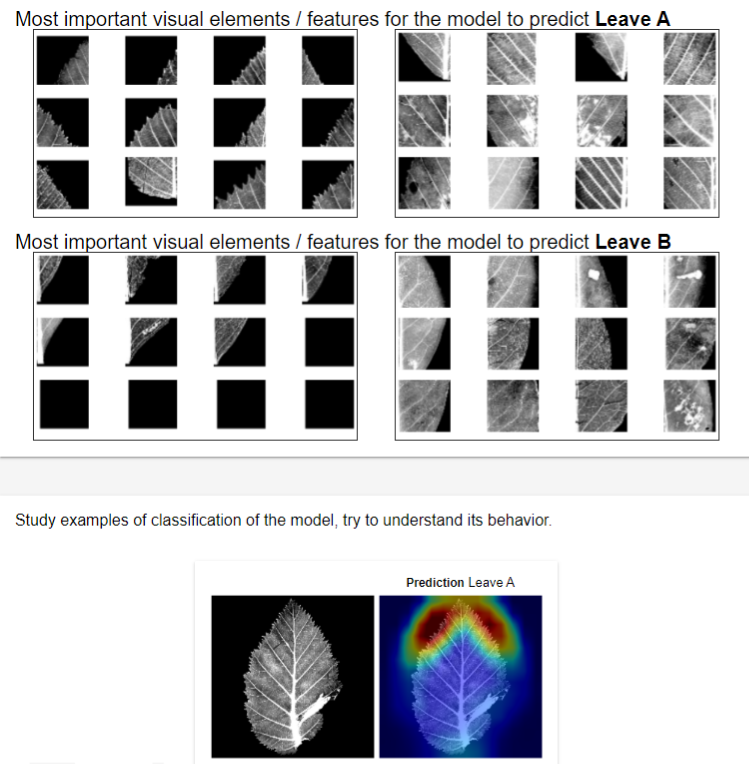}
        \caption{\textbf{Utility experiment.} Training trials taken from the \textit{Husky vs. Wolf} scenario (left) and the \textit{Leaves} scenario (right).}
        \label{fig:website_utility}
    \end{subfigure}
\end{figure*}

\subsection{Validation of Recursivity}

\paragraph{Participants} Behavioral accuracy data were gathered from $n=73$ participants. All participants provided informed consent electronically in order to perform the experiment ($\sim 4 - 6$ min). The protocol was approved by the University IRB and was carried out in accordance with the provisions of the World Medical Association Declaration of Helsinki. 
For each of the 2 experiment tested, we had prepared filtering criteria for uncooperative people (namely based on time), but all participants passed these filters.

\paragraph{General study design}

\begin{figure*}[hb]
    \centering
    \begin{subfigure}{0.95\textwidth}
        \includegraphics[width=\textwidth]{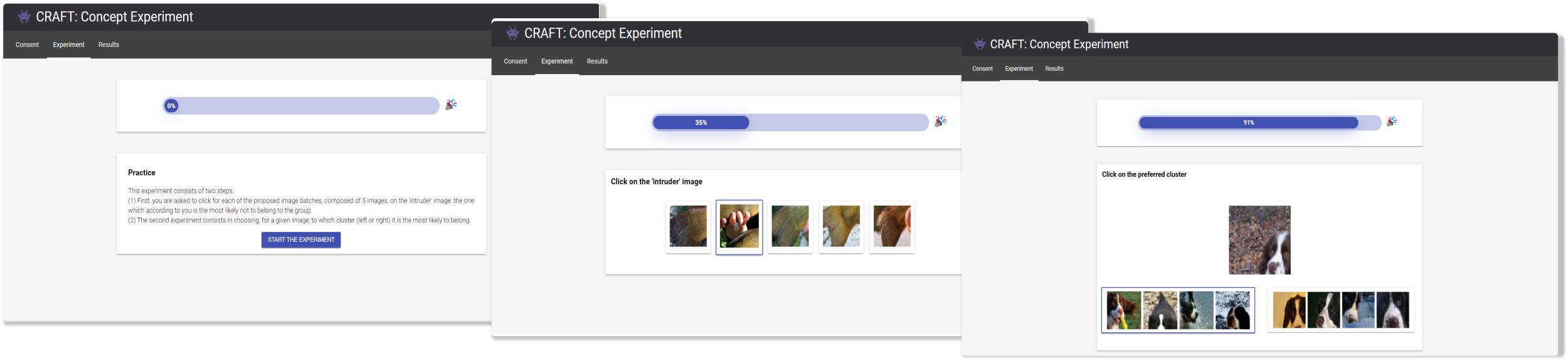}
        \caption{\textbf{Recursivity Experiment Website.}}
        \label{fig:website}
      
    \end{subfigure}
    \begin{subfigure}{0.95\textwidth}
        \centering
        \includegraphics[width=0.32\textwidth]{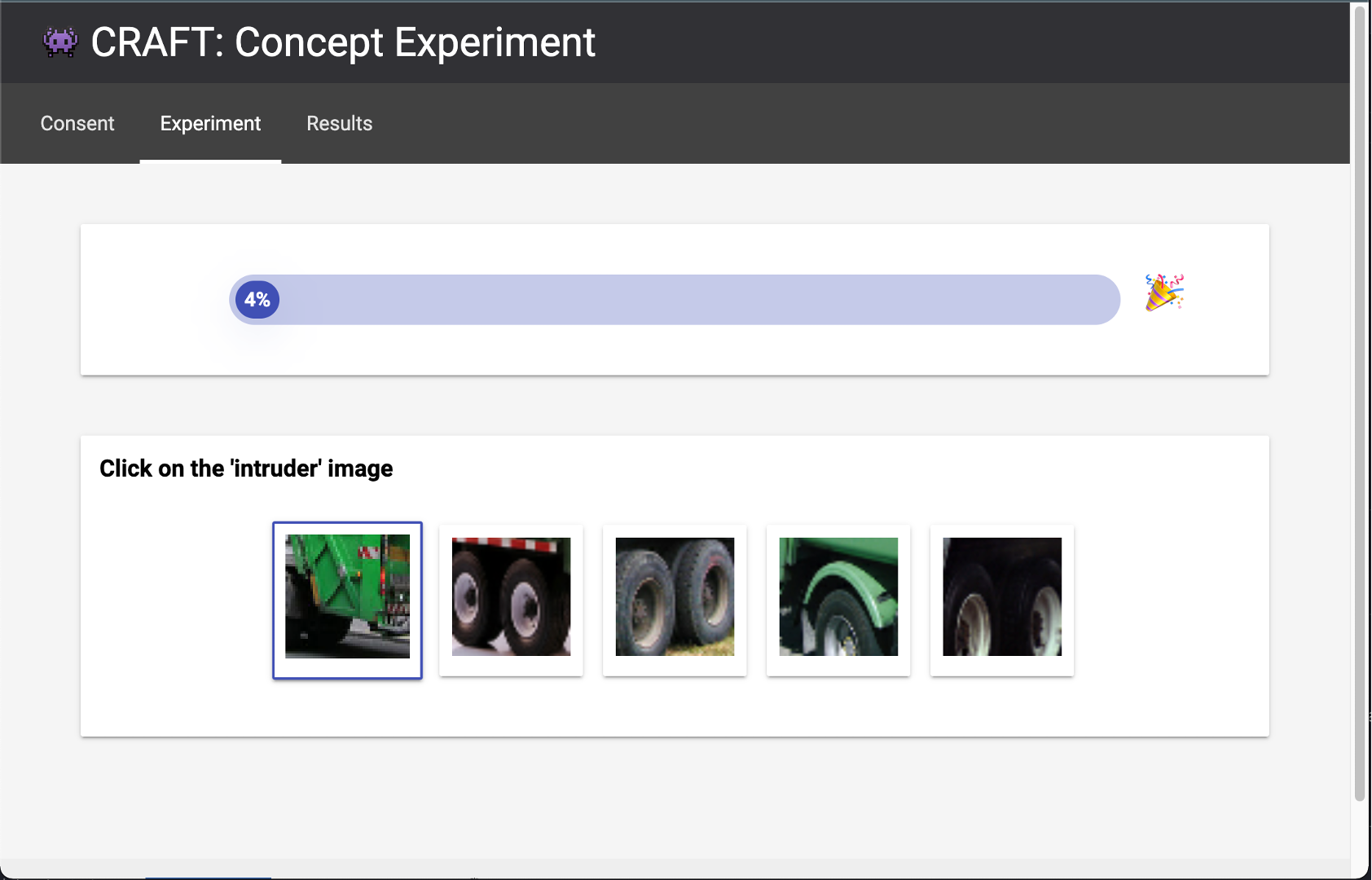}
        \includegraphics[width=0.32\textwidth]{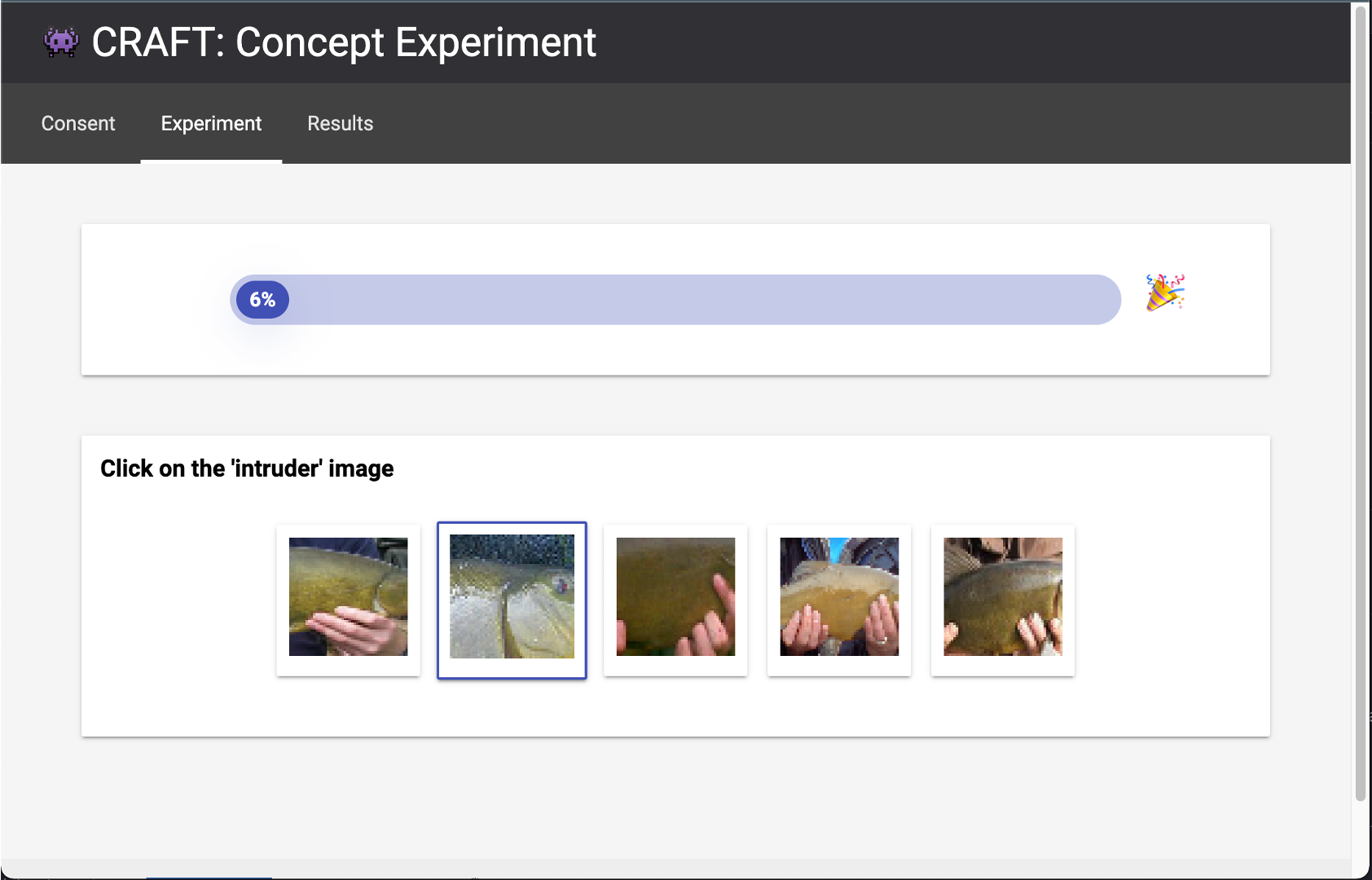}
        \includegraphics[width=0.32\textwidth]{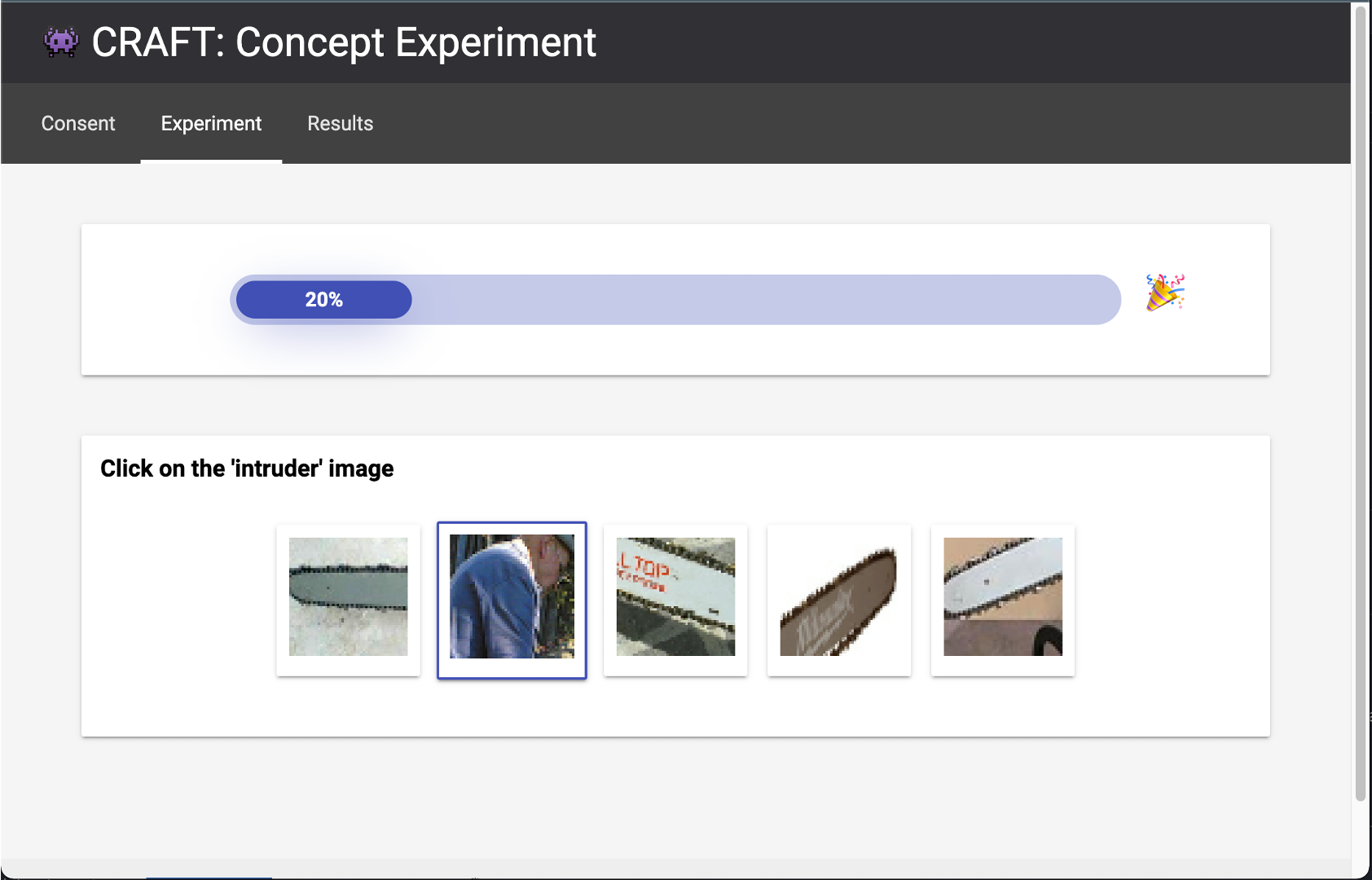}
        \caption{\textbf{Binary choice experiment.}}
        \label{fig:website_intruder}
    \end{subfigure}
    
    \begin{subfigure}{0.95\textwidth}
        \centering
        \includegraphics[width=0.32\textwidth]{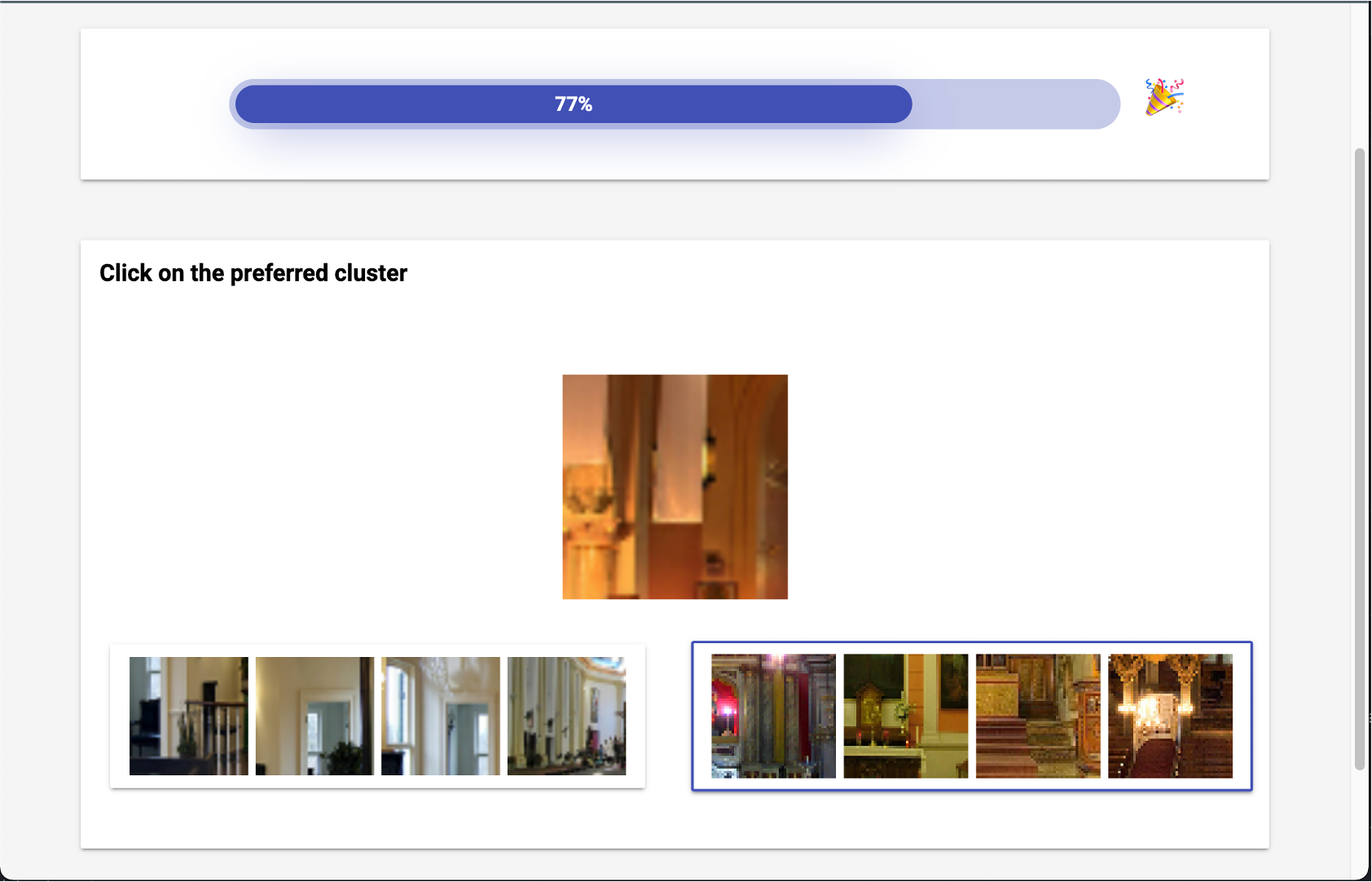}
        \includegraphics[width=0.32\textwidth]{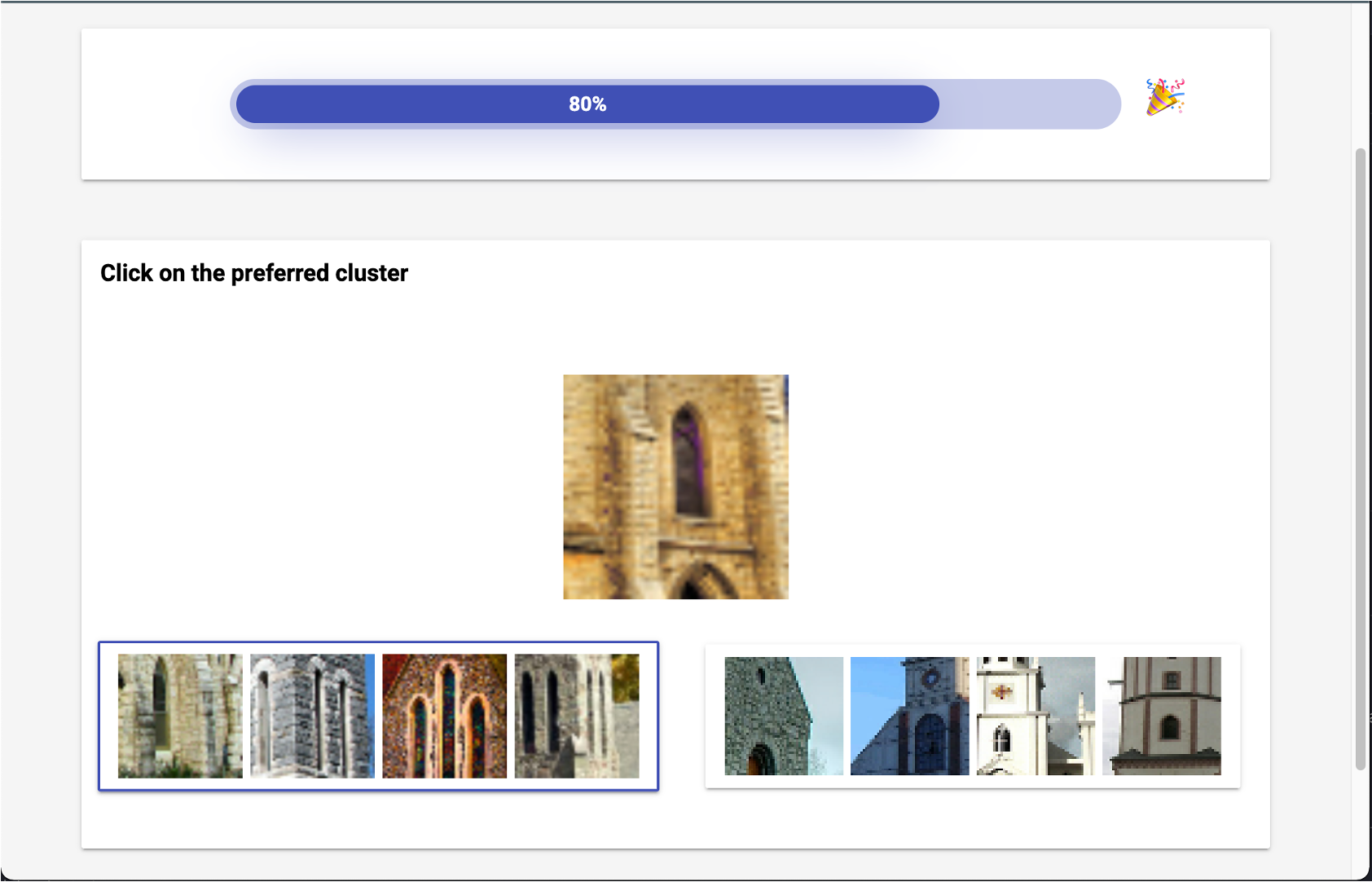}
        \includegraphics[width=0.32\textwidth]{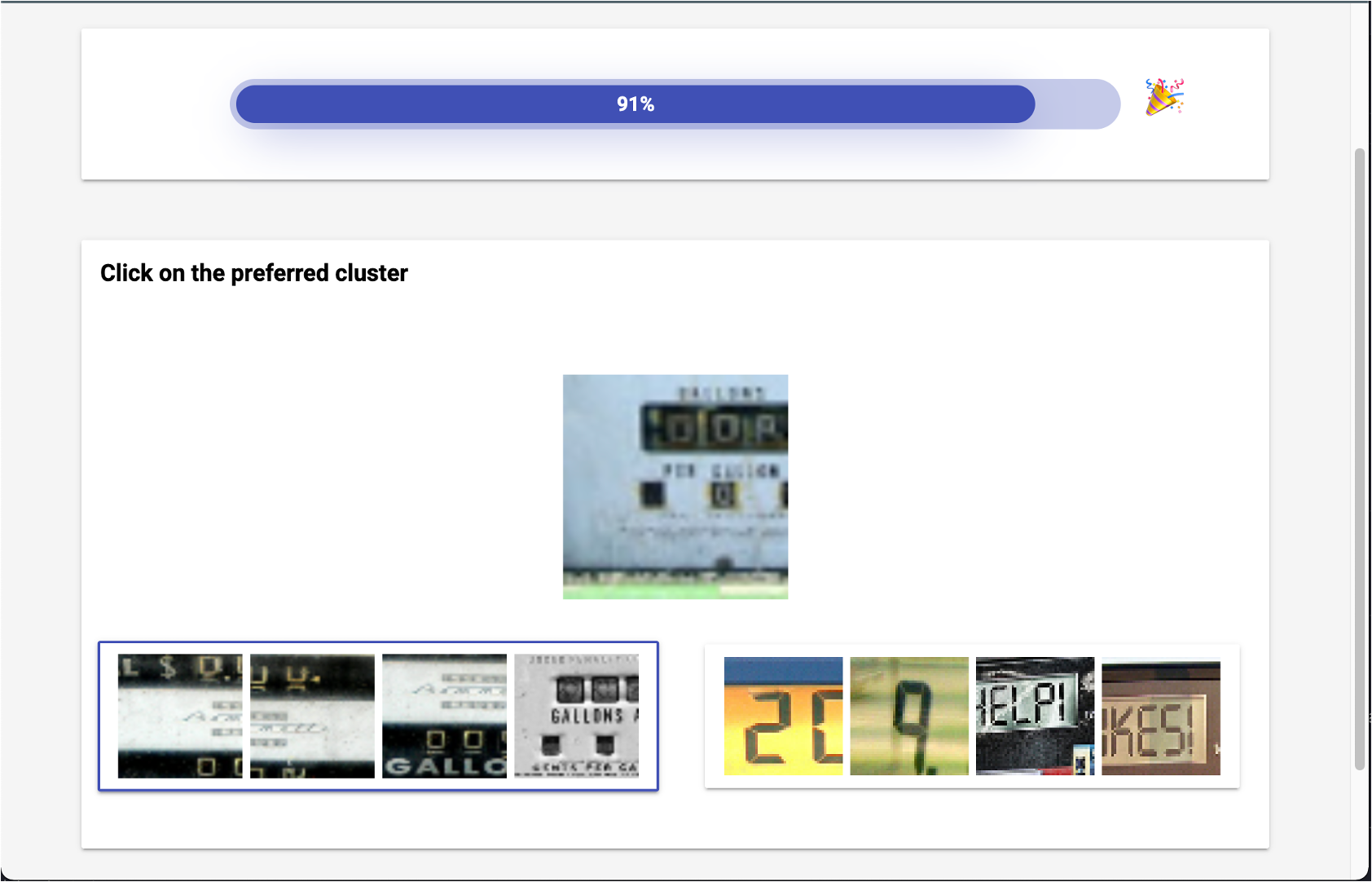}
        \caption{\textbf{Intruder experiment.}}
        \label{fig:website_choice}
    \end{subfigure}
\end{figure*}

For the first experiment -- consisting in finding the intruder among elements of the same concept and an element from a different concept (but of the same class, see Figure~\ref{fig:website_intruder}) -- the order of presentation is randomized across participants so that it does not bias the results.
Moreover, in order to avoid any bias coming from the participants themselves (one group being more successful than the other) all participants went through both conditions of finding intruders in batches of images coming from either concepts or sub-concepts.
Concerning experiment 2, the order was also randomized (see Figure~\ref{fig:website_choice}).

The participants had to successively find 30 intruders (15 block concepts and 15 block sub-concepts) for experiment 1 and then make 15 choices (sub-concept vs concept) for experiment 2, see Figure~\ref{fig:website}.

The expert participants are people working in machine learning (researchers, software developers, engineers) and have participated in the study following an announcement in the authors' laboratory/company. The other participants (Laymen) have no expertise in machine learning.

\clearpage

\section{Fidelity experiments}\label{app:fidelity}

\begin{minipage}{2.\columnwidth}

\begin{figure}[H]
  \includegraphics[width=.99\linewidth]{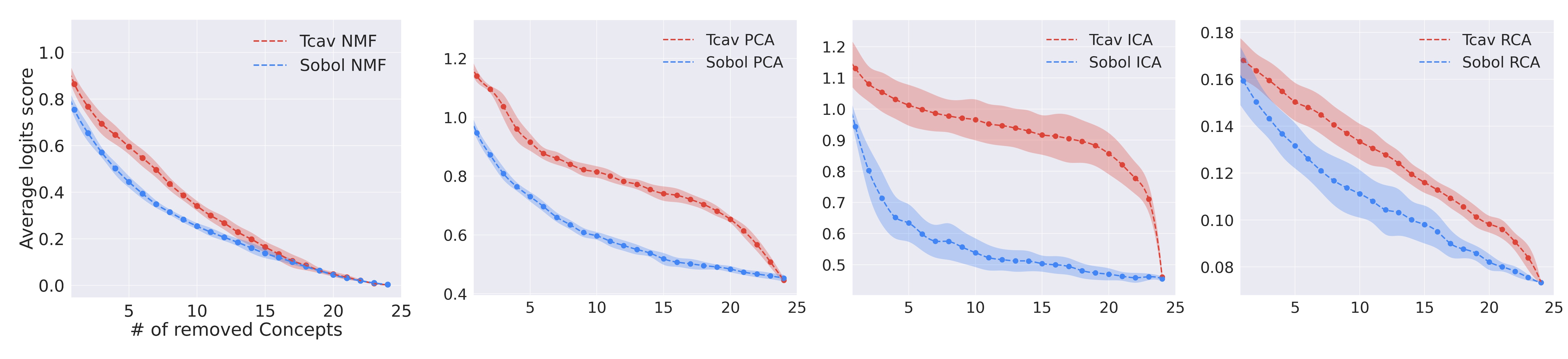}
  \includegraphics[width=.99\linewidth]{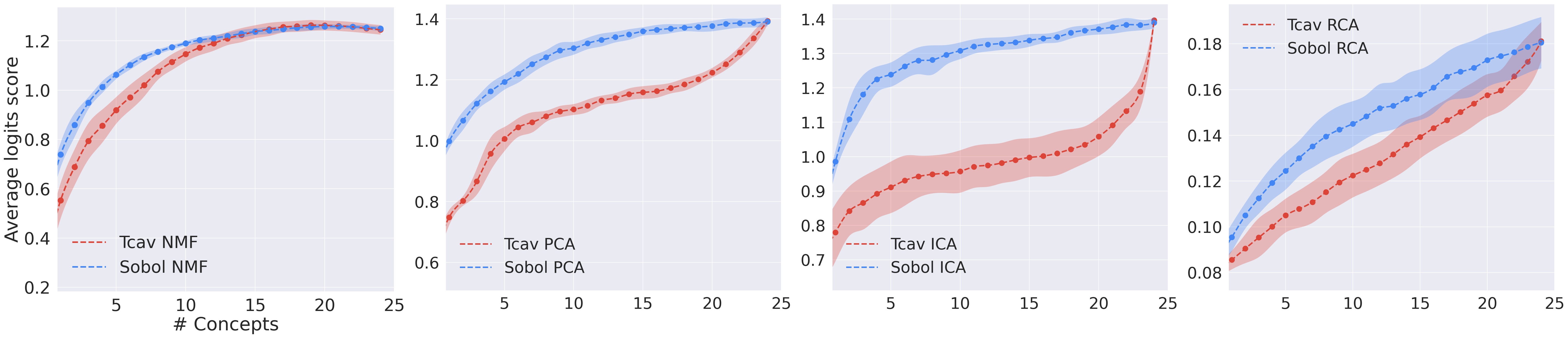}
  \caption{ \textbf{(1) Deletion curves} for different concept extraction methods, Sobol outperforms TCAV not only for NMF to correctly estimate concept importance (lower is better). \textbf{(2) Insertion curves} for different concept extraction methods, Sobol outperforms TCAV to correctly estimate concept importance (higher is better).}
  \label{fig:deletion_full}
\end{figure}

For our experiments on the concept importance measure, we focused on certain classes of ILSRVC2012~\cite{deng2009imagenet} and used a ResNet50V2~\cite{resnet50v2} that had already been trained on this dataset. Just like in~\cite{ACE, voleurs-didee-nmf}, we measure the insertion and deletion metrics for our concept extraction technique -- as well as concepts vectors extracted using PCA, ICA and RCA as dimensionality reduction algorithms, see Figure~\ref{fig:deletion_full} -- and we compare them when we add/remove the concepts as ranked by the TCAV score~\cite{TCAV} and by the Sobol importance score. As originally explained in~\cite{RISE}, the objective of these metrics is to add/remove parts of the input according to how much an explainability method considers that it is influential and looking at the speed at which the logit for the predicted class increases/decreases.

In particular, for our experimental evaluations, we have randomly chosen 100000 images from ILSVRC2012~\cite{deng2009imagenet} and computed the deletion and insertion metrics for 5 different seeds -- for a total of half a million images. In Figure~\ref{fig:deletion_full}, the shade around the curves represent the standard deviation over these 5 experiments.

\end{minipage}

\clearpage

\section{Sanity Check}
    \label{apx:sanity-checks}
        
\begin{figure}[h]
            \centering
            \includegraphics[width=0.95\linewidth]{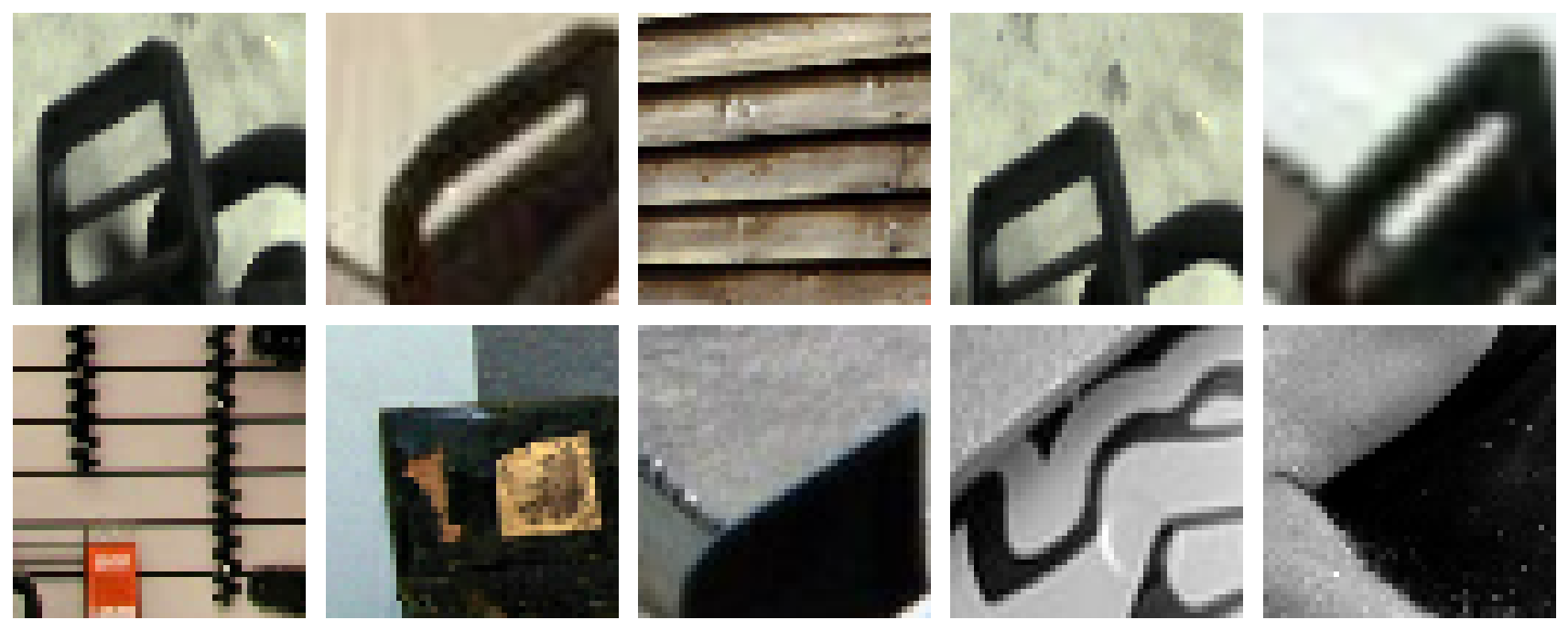}
            \includegraphics[width=0.95\linewidth]{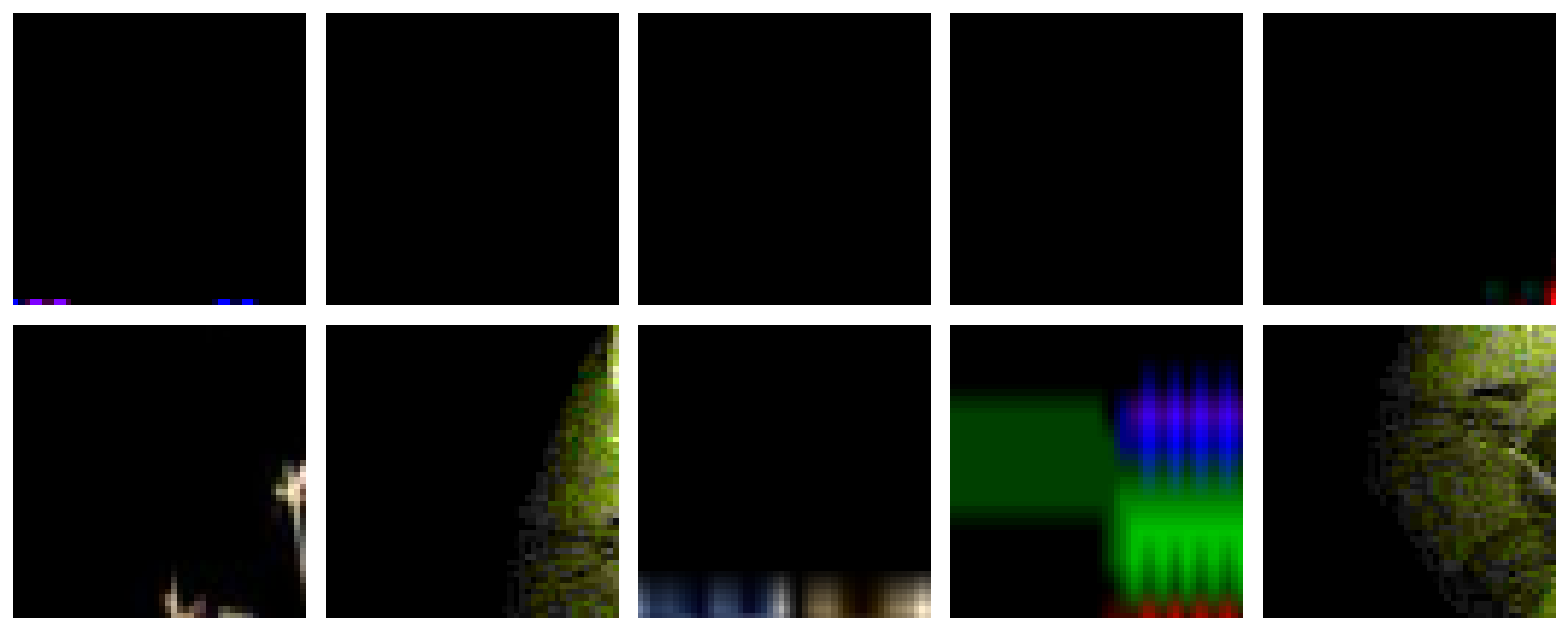}
            \includegraphics[width=0.95\linewidth]{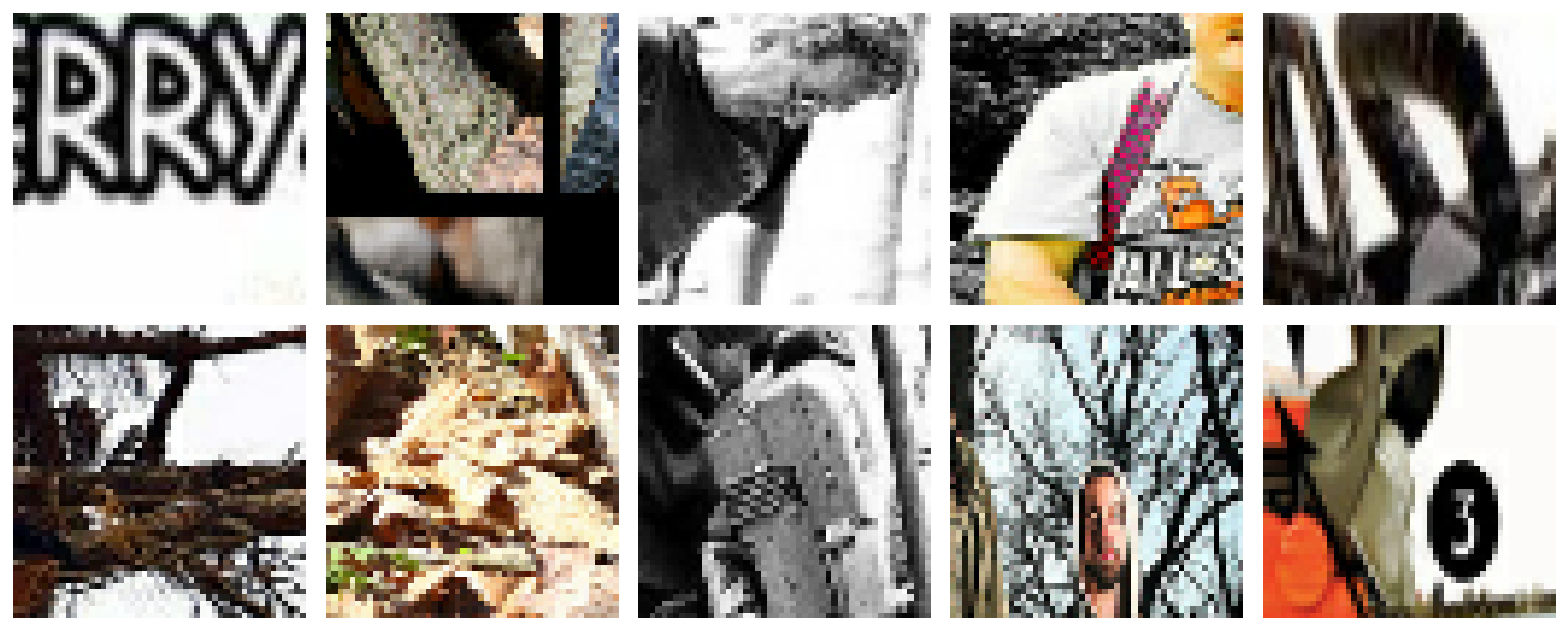}
            \caption{\textbf{Sanity check of the method:} we ran the method on a Resnet50 with randomized weights, and extracted the 3 most relevant concepts for the class `Chain saw'. When weights are randomized, concepts are mainly based on color histograms.}
            \label{fig:sanity_check}
        \end{figure}

        Following the work from~\cite{sanity-checks}, we performed a sanity check on our method, by running the concept extraction pipeline on a randomized model. This procedure was performed on a ResNet-50v2 model with randomized weights. As showcased in Figure~\ref{fig:sanity_check}, the concepts drastically differ from trained models, thus proving that CRAFT passes the sanity check.
        
\end{document}